%% file: acl_latex.tex
\documentclass[11pt]{article}

\usepackage[final]{acl}

\usepackage{times}
\usepackage{latexsym}
\usepackage{enumitem}
\usepackage{dblfloatfix}
\usepackage{booktabs}
\usepackage{multirow}
\usepackage{colortbl}
\usepackage{xcolor}
\usepackage{placeins}
\definecolor{oursblue}{RGB}{210,240,255}
\usepackage[T1]{fontenc}
\usepackage[utf8]{inputenc}

\usepackage{microtype}

\usepackage{inconsolata}

\usepackage{graphicx}
\usepackage{amsmath}
\usepackage{tabularx} 
\usepackage{float} 
\usepackage{makecell}

\title{PIVOTS-Bench: Evaluating Fine-Grained Interpersonal Relationship \\Reasoning in Multimodal Large Language Models
}

\author{
Shuxiang Zhang$^\diamond$ \\
Sun Yat-sen University \\
\texttt{zhangshx36@mail2.sysu.edu.cn} \\
\And
Yiting Yin$^\diamond$ \\
University of Michigan \\
\texttt{ciossayi@umich.edu} \\
\And
Wenxuan Song \\
Tsinghua University \\
\texttt{songwx24@mails.tsinghua.edu.cn} \\
\AND
Yuhang Wu$^\dagger$ \\
Tsinghua University \\
\texttt{yh-wu22@mails.tsinghua.edu.cn} \\
\And
Miao Liu\thanks{Corresponding author. \hspace{1em}
$^\diamond$Equal contribution. \hspace{1em}
$^\dagger$Project lead.} \\
Tsinghua University \\
\texttt{miaoliu@mail.tsinghua.edu.cn} \\
}
\begin{document}
\maketitle
\begin{abstract}
Humans possess an innate ability to understand fine-grained interpersonal relationships, which is central to everyday social interactions. Although such reasoning is inherently multimodal, it remains largely unexplored by existing multimodal large language models (MLLMs). To address this gap, we introduce \textbf{PIVOTS-Bench}, the first benchmark built from Social-IQ 2.0 and YouTube data to evaluate MLLMs’ ability to predict bidirectional interpersonal relationship dimensions grounded in established psychology research. In addition, PIVOTS-Bench includes auxiliary tasks that assess models’ ability to identify and leverage the critical visual cues underlying such predictions. We evaluate both proprietary and open-source MLLMs and conduct detailed ablation studies to analyze the effects of visual modalities and explicit social role information in conversational utterances. We further examine how joint and pairwise prediction settings benefit MLLMs in scoring bidirectional PIVOTS dimensions. \textbf{Project page:} \url{https://ciossayin.github.io/pivots-bench/}.
\end{abstract}

\begin{figure*}[ht]
    \centering
    \includegraphics[width=0.99\textwidth]{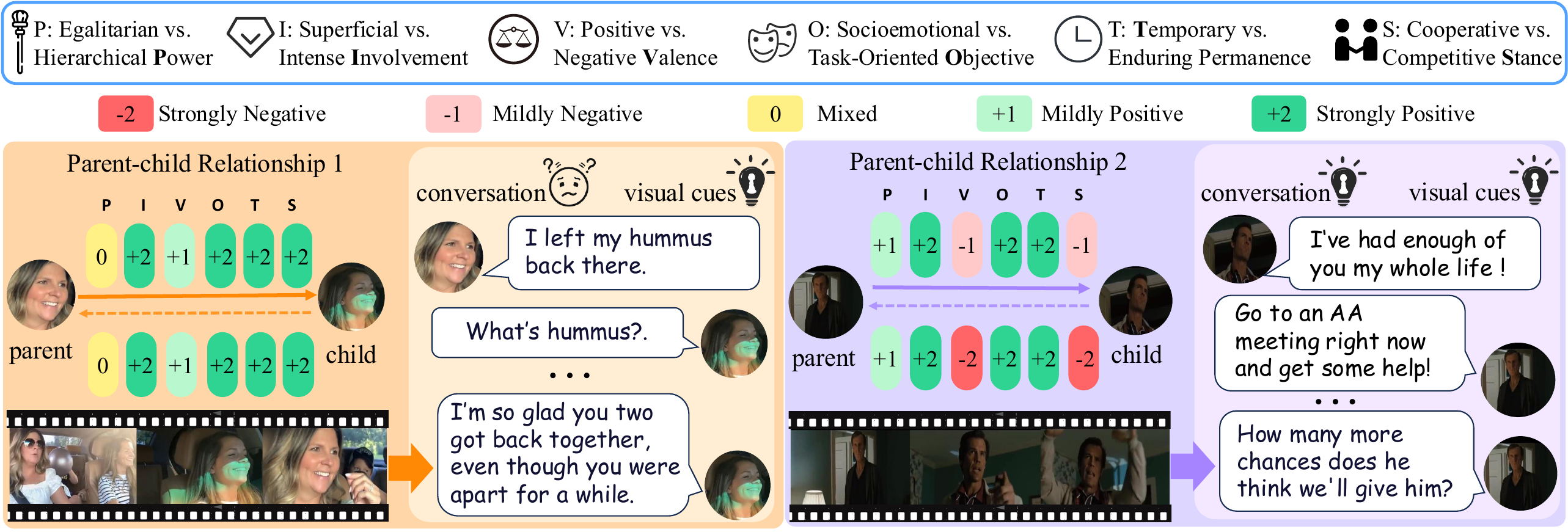}
    \caption{\textit{Understanding complex social dynamics requires joint reasoning over language and visual cues.} Although both examples involve parent–child interactions, they exhibit drastically different configurations across multiple interpersonal dimensions. Moreover, conversational utterances alone are often insufficient to reason about these dimensions, whereas visual cues provide complementary signals.}
    \label{fig:teaser}
\end{figure*}

\section{Introduction}

Humans routinely infer interpersonal relationships in everyday social interactions, a skill essential for collaboration, persuasion, trust formation, and conflict resolution. Moreover, our ability to attribute social relationships is inherently fine-grained and cannot be represented by conventional social categories such as family, friends, colleagues, or acquaintances. As illustrated in Fig.~\ref{fig:teaser}, although both examples depict a parent–child relationship, the underlying social dynamics differ dramatically: in the left example, the interaction is largely positive and cooperative, whereas in the right example it is tense and adversarial. An AI system capable of reasoning about such fine-grained interpersonal dimensions could both facilitate everyday human interactions and enhance multi-agent collaboration \cite{li2023camel} and socially aware decision making \cite{park2023generative}.

Recent advances in large language models (LLMs) have attracted interest in their potential for social intelligence~\cite{zhu2025social,mou2025agentsense,talebirad2025wisdom}, yet interpersonal relationship reasoning remains largely unexplored. Recent work has investigated whether existing LLMs can assess interpersonal relationships from text only corpora~\cite{zhou-etal-2025-socialeval}. It is worth noting that humans’ remarkable ability to infer social dynamics stems from the orchestration of multimodal signals. Using Fig.~\ref{fig:teaser} for illustration, the left example demonstrates a case where linguistic content alone is insufficient and visual cues are critical for interpretation, whereas in the right example visual and linguistic signals jointly inform social reasoning. Notably, a systematic study of whether Multimodal Large Language Models (MLLMs) can leverage multimodal cues to decipher social dynamics is still missing from the community.

To bridge this gap, we present PIVOTS-Bench, the first benchmark for evaluating MLLMs on interpersonal relationship reasoning, constructed using data curated from Social-IQ 2.0~\cite{siq2} and YouTube. Specifically, we consider six dimensions grounded in prior psychology research~\cite{wish1976perceived}: egalitarian vs.\ hierarchical power (\textbf{P}), superficial vs.\ intense involvement (\textbf{I}), positive vs.\ negative valence (\textbf{V}), socioemotional vs.\ task-oriented objective (\textbf{O}), temporary vs.\ enduring permanence (\textbf{T}), and cooperative vs.\ competitive stance. (\textbf{S})

Unlike prior work that models interpersonal relationships unidirectionally~\cite{rashid2017dimensions}, our benchmark adopts a \emph{bidirectional} setting that evaluates whether MLLMs can predict how each subject interprets the interpersonal relationship with the other across six dimensions. This design is motivated by information asymmetry, where two participants in the same interaction may form different beliefs about their relationship due to asymmetric perspectives, roles, and communicative intent. In addition to bidirectional dimension scoring, we introduce auxiliary visual reasoning tasks to evaluate whether MLLMs can identify and interpret the visual cues underlying interpersonal relationship predictions.

In our experiments, we first evaluate prevailing MLLMs on the PIVOTS benchmark, and then conduct a detailed analysis of how different input modalities and explicit role information in conversational utterances contribute to model performance. We further explore alternative prediction settings for interpersonal dimension scoring, showing that modeling multiple dimensions together and predicting bidirectional relationships in a paired manner can offer advantages over independent prediction in certain scenarios. Moreover, we investigate prompting strategies that inject human reasoning heuristics, examining how they yield clear improvements in salient social interactions while exhibiting limitations in contrastive and in-depth social scenarios.

Our main contributions are summarized as follows:
\begin{enumerate}
    \item We introduce \textsc{PIVOTS-Bench}, a new benchmark for evaluating multimodal large language models on interpersonal relationship reasoning.
    \item We conduct a comprehensive evaluation of existing MLLMs and analyze the effects of different input modalities, explicit role information, and prediction settings on performance.
\end{enumerate}

\section{Related Work}

\subsection{Interpersonal Relationship Understanding}
Social Relationship Recognition (SRR) has traditionally been formulated as a classification problem over a fixed taxonomy of relationship types \cite{sun2017domain,li2017dual,wang2010seeing,li2020visual,zhang2015learning,kim2025loversfriendsevaluatingllms}. Video based SRR methods typically focus on modeling temporal interaction structure. For example, MSTR introduced a multi scale spatio temporal reasoning framework and released the VISR benchmark \cite{liu2019social}. 
However, research in the social sciences~\cite{hinde1976interactions} suggests that human social interactions are highly complex and cannot be adequately captured by conventional social roles such as leader or teammate, as the same pair of roles may exhibit drastically different social dynamics across contexts and over time. More concretely, conventional social roles fail to capture the subtle, multimodal signals that actually drive social influence~\cite{hoogeboom2020closer}. Instead, fine grained interpersonal relationships, such as cooperation and competition, provide more informative signals for understanding social dynamics \cite{wish1976perceived}. This dimensional perspective has also been adopted in computational linguistics, where interpersonal relational dimensions are inferred from textual data \cite{rashid2017dimensions}. Closest to our motivation, \citet{kim2025loversfriendsevaluatingllms} introduce SCRIPTS, a text-only benchmark that labels relationship types at the dialogue level with uncertainty-aware annotations, using relational dimensions~\cite{wish1976perceived} only as auxiliary labels. However, its task remains undirected categorical prediction over text, whereas our benchmark treats dimensional configurations as the primary prediction target and models each relationship bidirectionally to capture asymmetric perceptions within a dyad. Importantly, relationship signals in real world interactions are often conveyed through multimodal behaviors rather than text alone, yet multimodal interpersonal relationship identification remains largely unexplored. To address this gap, we introduce a novel benchmark that evaluates whether MLLMs can infer six carefully defined interpersonal dimensions from videos and dialogues and reason about the visual cues underlying these relationships.

\subsection{Multi-modal Social Intelligence Benchmarks}
A rich body of literature has studied explicit verbal and nonverbal signal modeling in social settings, including tasks such as active speaker localization~\cite{grauman2022ego4d} and audiovisual diarization~\cite{xu2022ava}. In contrast, our investigation is more closely related to efforts aimed at developing computational models of social cognition. Early research \cite{sun2017domain,li2017dual} in social cognition primarily focused on Social Relationship Recognition (SRR) from static images and explored domain adaptation across diverse social scenes. More recently, video-based benchmarks such as MMToM-QA  \cite{jin2024mmtom} and AGENT \cite{shu2021agent} have utilized procedurally generated environments to probe core psychological reasoning and Theory of Mind.~\citet{lai-etal-2023-werewolf} introduce a multimodal benchmark that studies persuasion strategy modeling using both linguistic and nonverbal signals. Although numerous studies have explored general social intelligence~\cite{sap-etal-2019-social, siq2, mathur-etal-2025-social, xu2025socialmazebenchmarkevaluatingsocial, kong2025sivbenchvideobenchmarksocial, zhou-etal-2025-socialeval}, they lack a systematic framework for measuring interpersonal dimensions derived from social signals. Our PIVOTS benchmark addresses this gap by probing whether MLLMs can understand  pairwise interpersonal relationship dimensions, thereby pinpointing the strengths and limitations of AI systems on social cognition capability. We compare the core differences between our work and existing related works in Appendix~\ref{comp_related_work}.

\section{PIVOTS Benchmark}

\begin{figure*}[ht]
    \centering
    \includegraphics[width=0.99\textwidth]{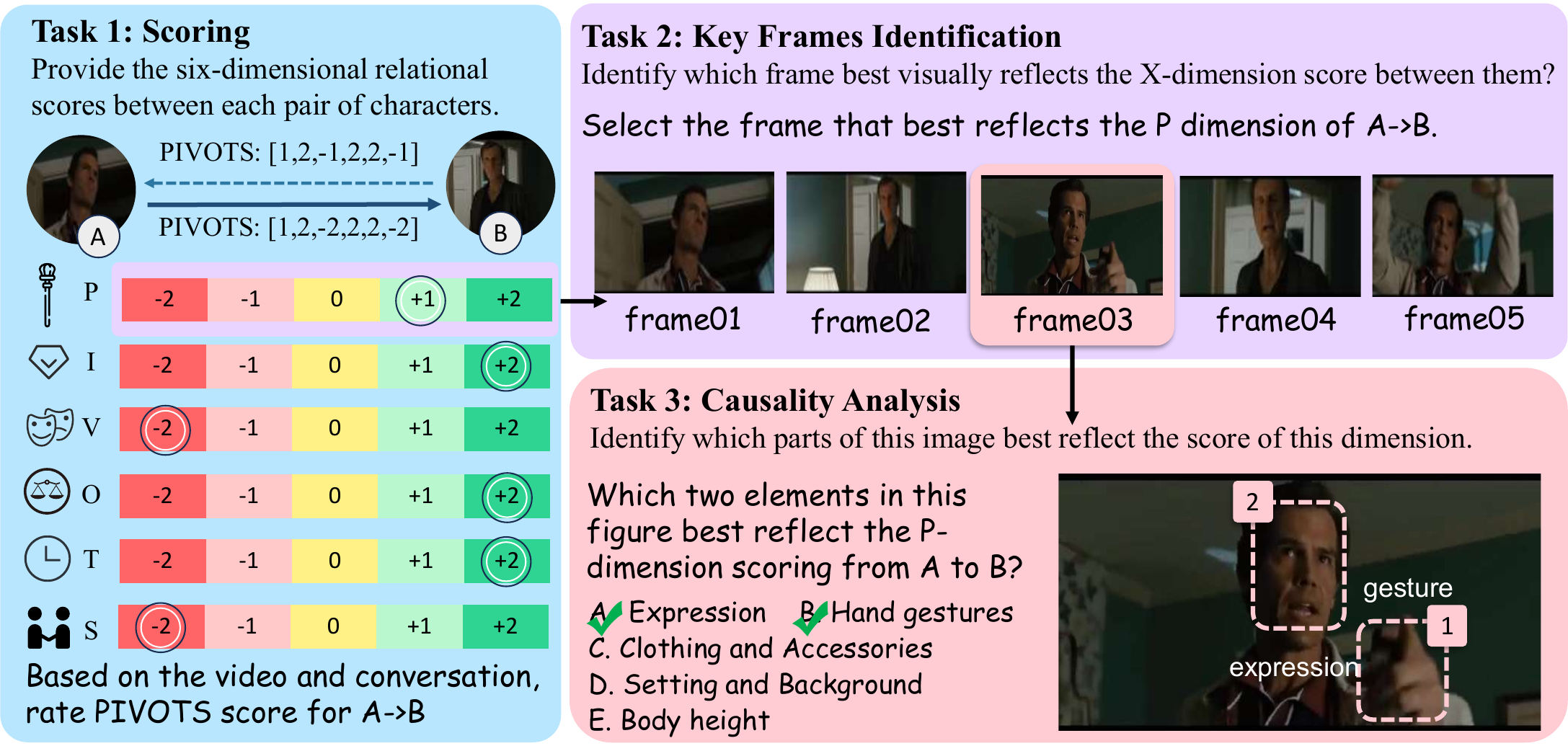}
    \caption{\textit{Overview of the three hierarchical tasks defined in the PIVOTS benchmark.} The scoring task asks models to assign six-dimensional PIVOTS relational scores from the video and conversation. The key-frame identification task requires selecting the frame that best reflects a specified relational dimension, while the causality analysis task reasons about the visual elements (e.g., facial expressions or gestures) that support the assigned score, enabling causal grounding of social judgments.}
    \label{fig:task}
\end{figure*}

\subsection{PIVOTS Dimensions Definition}
The foundation of our interpersonal relationship definition is grounded in the theoretical framework of~\citet{wish1976perceived}, which identifies four fundamental dimensions underlying interpersonal perception. We further expand the coverage by introducing two complementary axes, valence and temporal dimension, based on insights from prior research~\citep{russell1980circumplex}. We operationalize these six dimensions using a five-point bipolar Likert scale, ranging from -2 to +2. Formally, we consider the following six axes:

\begin{itemize}[leftmargin=*,nosep]
    \item \textit{Egalitarian vs. Hierarchical Power \big(P\big)}: This dimension relates to power distribution, status, and control, indicating whether the individuals involved have similar or different levels of power and roles.

    \item \textit{Superficial vs. Intense Involvement \big(I\big)}:
    This dimension reflects the depth, involvement, and emotional investment in the relationship.
    
    \item \textit{Positive vs. Negative Valence \big(V\big)}: This dimension captures the degree of pleasantness and satisfaction versus hostility in the relationship.
    
    \item \textit{Socioemotional vs. Task-Oriented Objective \big(O\big)}: This dimension distinguishes between relationships that are primarily personal and relaxed versus those that are goal-oriented, businesslike, and structured.
    
    \item \textit{Temporary vs. Enduring Permanence \big(T\big)}: This dimension distinguishes between relationships that are primarily long-term based and with specific kind of commitment or those that are short-termed and freewheeling.
    
    \item \textit{Cooperative vs. Competitive Stance \big(S\big)}: This dimension, also referred to as affiliation or affect, describes the degree of warmth, support, and collaboration in the relationship.
\end{itemize}

The goal of this dimensional system is precisely to quantify interactions with different surface forms on a unified, comparable scale, thereby moving beyond traditional coarse-grained categories such as parent-child, friends, colleagues. Detailed definitions, five-point rating rubrics for each dimension, and illustrative examples are provided in Appendix~\ref{sec:appen_definition}.

\subsection{Task Definition}
In this section, we formally define a set of benchmark tasks derived from the six interpersonal relationship axes introduced earlier. Apart from the primary task of scoring the PIVOTS interpersonal dimensions, we introduce auxiliary tasks for keyframe identification and grounded visual causal reasoning to assess models’ ability to utilize visual information effectively. We present a visual illustration of our tasks in Fig.~\ref{fig:task}.

\noindent\textbf{Task 1: Six-Dimensional Scoring}.

Given the video $\mathcal{V}$, conversation utterance $\mathcal{U}$, and a directional relationship $\mathcal{D}_{AB}^{\alpha}$, MLLMs aim to predict the corresponding score $\mathcal{R}_{AB}^{\alpha}$. Note that $\mathcal{D}_{AB}^{\alpha}$ specifies how subject A interprets subject B along one of the six interpersonal axes $\alpha \in \{P, I, V, O, T, S\}$. Specifically, we define the relationship score prediction via the following conditional probability:
\[
\mathcal{R}_{AB}^{\alpha}
=
\underset{r \in \{-2,-1,0,1,2\}}{\arg\max}
\ p\big(r \mid \mathcal{V}, \mathcal{U}, \mathcal{D}_{AB}^{\alpha}\big),
\]
\noindent where $r$ represents the discrete candidate score for each PIVOTS dimensions.

\noindent\textbf{Task 2: Key Frame Identification}.

Given video $\mathcal{V}$ and direction $\mathcal{D}_{AB}^{\alpha}$, MLLMs aim to identify the key video frame that best reflects the identification of the specified dimensional score. Specifically, we define this task via the following conditional probability:
\[
\mathcal{T}_{AB}^{\alpha} = 
\underset{t \in \{ I_1, I_2, \dots, I_n\}}{\arg\max}
\ p\big(t \mid \mathcal{V}, \mathcal{D}_{AB}^{\alpha}\big),
\]
where $\{ I_1, I_2, \dots, I_n\}$ is the candidate image sequence  sampled from video $\mathcal{V}$.

\noindent\textbf{Task 3: Visual Cue Causal Analysis}.

Given a key frame $\mathcal{T}_{AB}^{\alpha}$ and direction $\mathcal{D}_{AB}^{\alpha}$, MLLMs are prompted to select the visual cues that support the identification of the corresponding relationship dimension score. We formalize this task as the following conditional inference problem:
\begin{equation}
\mathcal{C}_{AB}^{\alpha} =
\underset{c \in \{ o_1, o_2, \dots, o_n\}}{\arg\max}
p \big(c \mid \mathcal{T}_{AB}^{\alpha}, \mathcal{D}_{AB}^{\alpha}\big),
\end{equation}
where $\{o_1, o_2, \dots, o_n\}$ denotes a set of candidate visual cues extracted from the key frame.
For a given query $\mathcal{Q}_{AB}^{\alpha}$, the selected cue $\mathcal{C}_{AB}^{\alpha}$ corresponds to the image evidence (e.g., facial expressions, hand gestures, or body posture) that supports the identification of $\mathcal{R}_{AB}^{\alpha}$.

\subsection{Data and Annotation}
\label{sec:data_main}
\noindent\textbf{Data Source}.\ To systematically study the tasks defined in our benchmark, we design our data curation strategy to prioritize scenarios that show prominent social signals and rich dynamics of interaction. We derive our dataset from two primary sources:

\noindent\textit{Social-IQ 2.0:}

We construct a high-quality subset of 121 videos based on the Social-IQ 2.0 dataset \cite{siq2}, focusing on explicit social interaction scenarios. To strictly mitigate visual biases and information leakage risks, we manually eliminate samples involving news interviews and explicit visual markers (e.g., text overlays), while uniformly segmenting all selected videos into 30-second clips. The final dataset accurately captures interaction features such as debates and informal conversations, providing sufficient support for multi-dimensional fine-grained social relationship analysis.

\noindent\textit{YouTube Videos:} Note that existing work on social relationship understanding often focuses on overt and expressive social behaviors, whereas many real-world spontaneous interactions are more subtle, restrained, and socially nuanced. To mitigate this limitation, we augment our corpus with YouTube data, structured across two primary dimensions:
\begin{itemize}[leftmargin=*,nosep]
    \item Interaction Dynamics: We target keywords associated with complex communicative goals, such as persuasion, and conflict resolution.
    \item Relational Contexts: The dataset spans a spectrum of social bonds, from intimate dyads (e.g., romantic and dissolved relationships) and familial ties (e.g., siblings) to zero-acquaintance dynamics among participants with divergent ideological backgrounds. 
\end{itemize}
\noindent 
Notably, the interpersonal relationships in this curated subset are primarily reflected by the overall interaction atmosphere rather than specific body gestures within particular temporal segments. As a result, this subset is less suitable for keyframe identification or fine-grained causal analysis of individual visual cues. We make this trade-off to diversify the coverage of interpersonal relationship understanding. This additional data source comprises 70 videos and effectively complements the predominantly casual interactions in Social-IQ 2.0.

\noindent\textbf{Data Selection}.\ Following the grounding principles, we filtered the raw collection to ensure multi-modal alignment. Specifically, we required that: (1) multiple speakers are visible and audible; (2) the interaction lasts longer than 30 seconds to allow for relationship development; and (3) the social dynamic is not interpretable with trivial symbols.

\noindent\textbf{Annotations}.\
We adopted a unified “model-first, human-calibrated” annotation framework across both video sources. Concretely, we consolidated the reference evidence, basic video metadata, and the PIVOTS six-dimensional definitions with their rating guidelines into a single prompt, and first used the GPT-5 API to generate initial per-character scores across all six dimensions. We then treated these model-generated ratings as a starting point and performed systematic human annotation, ensuring that each score strictly aligns with the dimension definitions. According to our statistics, the proportion of scores we revised based on the GPT-5 initial outputs is 25.3\%. We conducted a validation study on 35 videos to examine whether model-assisted annotation introduces systematic bias. The agreement between fully human annotations and model-assisted annotations on this subset is 0.8399~\cite{1960A}. We treat this as a preliminary consistency check rather than as evidence that the workflow is free of bias, and discuss its limitations in the Limitations section. Additional annotation details are provided in Appendix~\ref{sec:annot_details}.

\noindent\textbf{Inter-Annotator Agreement}.\
We compute inter-annotator agreement using the ordinal form of Krippendorff’s alpha~\citep{krippendorff2018content}. The overall Krippendorff’s $\alpha$ value is 0.8125, indicating that the dataset annotations are reliable. We refer to Appendix~\ref{sec:annot_agree} for additional details.

\input{latex/tables/overall}

\noindent\textbf{Benchmark Statistics.}

\begin{figure}[!h]
\centering
\includegraphics[width=1.0\linewidth, page=2]{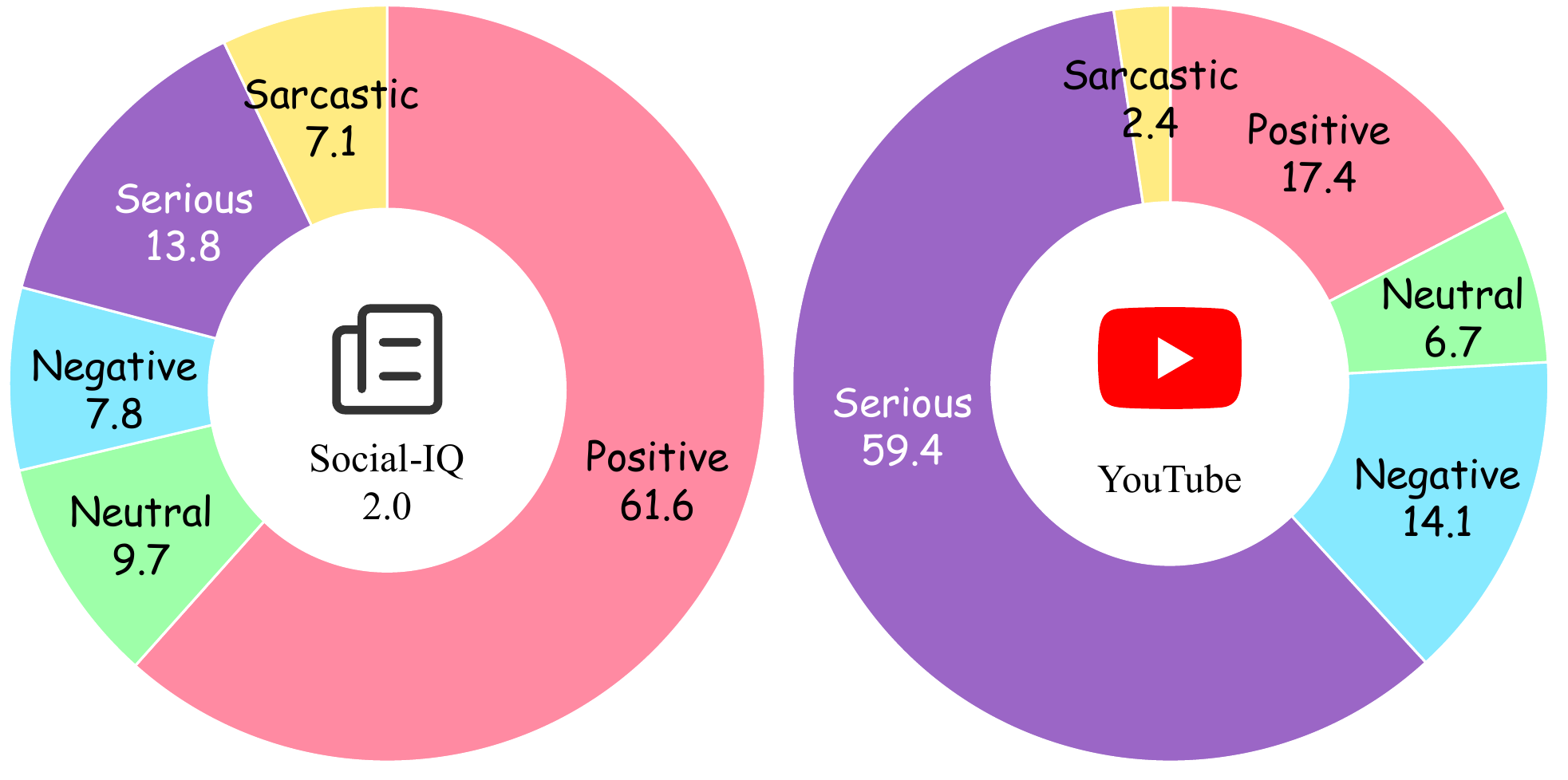}
\caption{Label distributions across emotion and tone dimensions for Social-IQ 2.0 and YouTube videos.}
\label{fig:tone}
\end{figure}

Fig.~\ref{fig:tone} presents the statistical distribution of the Social-IQ 2.0 data and the YouTube data. Notably, the distributions across the two data sources differ substantially: positive is the dominant category in Social-IQ 2.0, whereas serious is the primary category in the YouTube data. This difference aligns with our earlier discussion, as we intentionally source more in-depth interactions from YouTube to complement and diversify the benchmark. A detailed comparison of the two datasets is presented in Appendix~\ref{sec:comparison between}.

Our final benchmark comprises 595 VQA pairs per PIVOTS dimension from Social-IQ 2.0 and 170 per dimension from YouTube. We further curate 483 additional VQA pairs for auxiliary tasks of key frame identification and visual cue causal analysis. Detailed statistics are presented in Appendix~\ref{sec:detailed statistics}.

\section{Experiments}
\subsection{Experiment Setup and Metrics}
We evaluate both open-source and proprietary MLLMs on our benchmark, including Gemini-2.5~\cite{comanici2025gemini}, GPT-5~\cite{openai2025gpt5}, and Qwen-3 models~\cite{qwen3technicalreport} of varying sizes. For input-modality ablations and additional analyses, we adopt the strong open-source model Qwen-32B-VL as the base model.

The scores of PIVOTS axes are defined on a five-point discrete scale from $-2$ to $+2$, yielding a chance-level accuracy of 20\%. The key-frame identification and causal analysis tasks are formulated as five-option multiple-choice questions, which likewise result in a chance-level accuracy of 20\%. For visual cue causal analysis, models are required to select the two most relevant cues from five candidates. Partial credit of 0.5 is awarded when one selected cue is correct, while full credit of 1.0 is given when both selected cues are correct. Under random selection, the expected chance-level accuracy is 40\%.

\subsection{Experimental Results}

\noindent\textbf{Benchmark Performance}.\
We report the experimental results of prevailing MLLMs on our benchmark in Tab.~\ref{tab:social_iq} and Tab.~\ref{tab:youtube}. In terms of PIVOTS dimensions scoring, proprietary models achieve substantially stronger performance across all dimensions on both Social-IQ 2.0 and YouTube. Although open-source models often achieve on-par performance with proprietary models on prevailing MLLM benchmarks, they still struggle with fine-grained social dynamics reasoning, which motivates the need for benchmarks that evaluate this capability directly.

We also observe that MLLMs achieve better performance on the YouTube subset than on the Social-IQ 2.0 subset. One possible explanation is that YouTube content is widely available online and may have been more extensively represented in model pretraining than the licensed film clips underlying Social-IQ 2.0. We note, however, that this form of exposure does not amount to label leakage: the six PIVOTS dimensions are annotated by us and have not been publicly released, so a model that has encountered a video during pretraining has not thereby encountered the directional scores it is asked to predict. What source familiarity may confer is an advantage in scene comprehension rather than knowledge of the relational judgments themselves. Two observations are consistent with this reading. First, the size of the gap varies across models, ranging from under two points for GPT-5 to over seven for Qwen3-8B-vl. Second, prediction settings and prompting strategies behave in opposite directions on the two subsets: joint and pairwise prediction improve performance on Social-IQ 2.0 but degrade it on YouTube (Table 5), and both prompting strategies show the same reversal (Table 6). Familiarity with the source material would not predict this pattern.

Among the six axes, we find that the valence-v and objective-o dimensions are notably more challenging for all models. The difficulty of valence prediction is exacerbated by its reliance on subtle affective cues such as tone, which is not captured by most current multimodal large language models. And objective judgments become unreliable when task-driven interactions are accompanied by strong emotional expressiveness.

For Task 2 and Task 3, Gemini and GPT also consistently outperform Qwen3. In particular, on the visual cue causal analysis task, Qwen3 often performs below the chance-level baseline, indicating difficulty in identifying the relevant visual cues required for reasoning about interpersonal relationship dimensions. This result suggests a notable limitation in the Qwen series’ ability to perform social behavior–related visual reasoning.

\input{latex/tables/modality}
\input{latex/tables/audio}
\input{latex/tables/setting}
\input{latex/tables/withcot}

\noindent\textbf{Ablation on Input Modalities}.\
We conduct detailed ablation studies in Tab.~\ref{tab:modality} to analyze how different input modality configurations and text redaction settings affect multimodal interpersonal relationship reasoning.

It is possible that models leverage explicit social relationship identifiers as shortcuts for inferring interpersonal relationships. Such identifiers refer to textual components in the original utterances that directly reveal participants’ identities or relational attributes, including but not limited to kinship terms (e.g., father, daughter) and professional titles or honorifics (e.g., Your Honor, boss). In this context, we consider a baseline that uses redacted conversation utterances as inputs, in which explicit social identifiers are replaced with generic tokens that contain no social attribute information (e.g., [PERSON], [TITLE]), while preserving the syntactic structure, emotional tone, and logical flow of the conversation. As shown in Tab.~\ref{tab:modality}, redacting explicit social identifiers results in a performance decrease of 2.47\% on Social-IQ 2.0 and 4.17\% on the YouTube subset relative to using the original utterances. This result suggests that explicit social roles provide useful cues for scoring PIVOTS dimensions. However, it also indicates that models do not rely solely on such identifiers as shortcuts for making predictions.

We further assess the contribution of visual modalities. On the Social-IQ 2.0 set, incorporating video information improves performance by 3.1\% over the original utterance baseline and 6.86\% over the redacted utterance baseline. This suggests that visual modalities contribute more strongly to interpersonal reasoning in Social-IQ 2.0 scenarios with salient social behaviors, especially in the absence of explicit social role cues. For the YouTube subset, incorporating video input yields a 3.0\% performance gain when using the original utterances, but results in a 1.42\% performance decrease when combined with redacted utterances. Meanwhile, we use Qwen3-Omni-30B to evaluate the contribution of the audio modality (this setup utilizes original transcriptions and incorporates visual inputs). The results are shown in Table~\ref{tab:audio}. Upon ablating the audio modality, its performance drops by 8.32\% on the Social-IQ 2.0 dataset, while it only decreases by 1.80\% on the YouTube dataset. The divergence can be attributed to how we curated the YouTube subset. As discussed in Sec.~\ref{sec:data_main}, these videos primarily capture in-depth, restrained interactions in which interpersonal relationships are conveyed through global discourse context and shared background rather than salient, localized visual cues. When explicit role information is removed, models struggle to anchor visual observations to participants’ identities and intentions, making it difficult to integrate visual signals with conversational context for reliable interpersonal state inference.

\subsection{Additional Analysis}
\noindent\textbf{Prediction Setting}.\ We consider three different prediction settings for scoring each interpersonal relationship dimension:
\begin{itemize}[leftmargin=*,nosep]
    \item \textbf{Separate prediction:} This is the baseline setting that scores each PIVOTS dimension independently, treating the perception of one subject toward another along a single dimension as an individual question.
    \item \textbf{Joint prediction:} The model is prompted to jointly rate six PIVOTS dimensions for one subject’s perception of another.
    \item \textbf{Pairwise prediction:} The model is required to predict all bidirectional PIVOTS dimension scores between two subjects.
\end{itemize}
The exact prompts used for each prediction setting are provided in Appendix~\ref{sec:prompt_infer}.

The experiments are conducted with videos and original utterances as inputs, and the results are summarized in Table \ref{tab:setting}. Notably, the effectiveness of prediction settings exhibits dataset-specific differences: On the Social-IQ 2.0 dataset, which primarily features casual, everyday interactions, joint prediction achieves the best performance (Avg 43.26\%), followed by pairwise prediction (42.19\%), both significantly outperforming separate prediction (34.54\%). This aligns with the mechanism of human social perception, as joint prediction encourages the model to reason based on the shared latent representation of interactions, enabling cues from different dimensions to complement each other and enhance the coherence and accuracy of social dynamics understanding.

In contrast, an opposite trend is observed on the YouTube dataset, which contains in-depth interactions such as in-depth interviews: Separate prediction yields the optimal average performance (40.81\%), outperforming joint prediction (38.63\%) and pairwise prediction (39.67\%). However, pairwise prediction performs exceptionally well on the P-dimension (86.96\%), while separate prediction maintains the best performance across dimensions including I, O, T, and S. This indicates that in in-depth social interactions, holistic modeling (joint/pairwise prediction) may interfere with the model's judgments on specific dimensions, whereas independent scoring is more conducive to capturing the unique cues of each dimension.

\noindent\textbf{Heuristic Injection via Prompting}.\ As evidenced by the experimental results in the preceding sections, existing MLLMs, particularly open-source models, often struggle to accurately score the PIVOTS dimensions. We therefore investigate how interpersonal relationship reasoning can be enhanced by drawing inspiration from human reasoning strategies. Concretely, we inject human reasoning processes into the prompt to guide MLLMs toward more accurate predictions. Specifically, we consider the following two prompting strategies:

\begin{itemize}[leftmargin=*,nosep]
    \item \textbf{Multi-Stage Prompting} first calibrates affective judgments (valence and intensity), then evaluates structural interpersonal dimensions (power, objective, permanence, and stance), followed by independent synthesis and reconciliation to ensure reliable PIVOTS predictions.
    \item \textbf{In-Context Prompting} grounds reasoning in fine-grained behavioral and contextual cues while explicitly modeling directional asymmetry between interacting individuals.
\end{itemize}
The exact prompts are provided in Appendix~\ref{sec:prompt_infer}.

The results of different prompting strategies are shown in Table \ref{tab:prompt}. On Social-IQ 2.0, multi-stage and context-aware prompting improved overall performance by 6.69\% and 5.97\% respectively, with the gains primarily stemming from improvements in the Involvement and Valence dimensions. However, for dimensions involving asymmetric perspectives, such as Objective, Power, and Stance, the improvement was limited, suggesting that prompting strategies failed to address the key bottleneck of aligning the correct social participants in complex interactions.

On the YouTube dataset, we observed a contrasting trend compared to Social-IQ 2.0. Both prompting strategies resulted in a significant decline in overall performance, with the Power dimension (for multi-stage prompting) and Objective/Stance dimensions (for in-context prompting) suffering the largest drop. This suggests that in deep and conservative social interactions, relying on rigid calibration or specific cues may instead confuse the models. This result highlights the necessity of supplementing the dataset with YouTube data; its diverse interaction settings drive research into computational social AI models that avoid overfitting to only salient and explicit interaction cues.
\section{Conclusions}
In this work, we introduce PIVOTS, the first benchmark for evaluating multimodal large language models on interpersonal relationship reasoning, together with auxiliary tasks that assess visual cue understanding. We first benchmark several prevailing MLLMs and then provide a detailed analysis of input modalities, prediction settings, and prompting strategies. Our experimental results show that explicit social scenarios and in-depth social scenarios pose significantly different challenges for computational social intelligence models. We believe our work provides a foundational step toward understanding fine-grained social dynamics and points to future research directions for designing robust and generalizable social AI models. 

\section*{Limitations}

Although our benchmark introduces the first fine-grained multimodal benchmark for interpersonal relationship reasoning, several limitations remain.

\noindent\textbf{Annotation was performed entirely by the authors}.\ All PIVOTS labels were produced by members of the research team rather than by external or crowdsourced annotators. While this allowed annotators to internalize the six-dimensional rubric in depth, which is difficult to achieve with short crowdworker training, it also means that
the annotators were aware of the goals of the study. We report inter-annotator agreement in Appendix~E, but we cannot rule out shared assumptions among annotators who developed the framework together. Validation by annotators fully external to the project remains important future work.

\noindent\textbf{The model-first pipeline may induce anchoring effects that our validation cannot fully exclude}.\ Initial scores were generated by GPT-5 and subsequently reviewed and corrected by human annotators, with 25.3\% of scores revised. Our validation study compares fully human annotations against model-assisted annotations on 35 videos and finds high agreement, but high agreement is consistent both with an unbiased workflow and with annotators anchoring on model outputs, and it therefore does not by itself establish the absence of bias. The validation subset is also small relative to the 191 videos in the benchmark. This concern is compounded by the fact that GPT-5 is itself among the evaluated models: its benchmark scores may be partially inflated by agreement with priors introduced during pre-annotation, and we did not isolate this effect in the present version. Comparisons involving GPT-5 should be interpreted with this in mind; the other evaluated models did not participate in annotation and are unaffected. Fully independent human annotation at scale is left to future work.

\noindent\textbf{Pretraining exposure to the source videos cannot be ruled out}.\ A fully contamination-controlled evaluation would require held-out material postdating the models' training cutoffs, which we were unable to construct for this version. We discuss the extent to which such exposure is likely to affect our results in Section~4.2.

\noindent\textbf{The benchmark is limited to English-language data}.\ This constrains its applicability to multilingual and cross-lingual social reasoning settings.

\noindent\textbf{Annotators share similar cultural and linguistic backgrounds}.\ None are native English speakers, which may introduce cultural bias in the interpretation of interpersonal relationships, particularly for the more subjective dimensions.

\noindent\textbf{We do not investigate instruction-tuning or reinforcement-learning approaches}.\ These methods require high-quality supervision that remains costly and difficult to scale, which is precisely the constraint that limits the size of our benchmark. Designing scalable data engines for obtaining reliable social reasoning supervision is an important direction for future work, and our benchmark provides a foundation for such efforts.

\section*{Ethical Considerations}
Notably, our benchmark involves the analysis of human subjects and therefore raises potential privacy concerns. For data sourced from Social-IQ 2.0, we follow the original dataset license, which permits use for research purposes. For YouTube videos, we adopt ethical practices consistent with prior work by releasing only video identifiers and temporal annotations, rather than distributing raw video content. We are committed to respecting individual privacy and ensuring compliance with established ethical standards in dataset creation, release, and usage. In addition, because our benchmark focuses on interpersonal relationship understanding, care must be taken to prevent potential misuse, such as deception, biased interpretations, or emotional manipulation. To mitigate these risks, we frame the benchmark as an analysis and evaluation tool rather than a deployable system, provide clear documentation on intended use and limitations, and discourage deployment in high-stakes or manipulative settings without human oversight.

\section*{Acknowledgments}
This work was supported in part by the Zhiyuan Scholar Program of the Beijing Municipal Science and Technology Commission (Z251100008125045), NSFC Grants, and a research grant from the ByteDance Seed Team.

\bibliography{custom}

\appendix

\section{Analysis and Discussion}
\subsection{Comparison with Existing Benchmarks}
\label{comp_related_work}

\input{latex/tables/related_work}
To present a comparative analysis with existing related benchmarks, we compare our work with five related works in Table~\ref{tab:related_work} . We perform an analytical comparison across four dimensions: sample size, core tasks, data sources, and annotation types. Compared with similar benchmarks, PIVOTS-Bench exhibits targetedness and sophistication in both sample size and task design. Our six-dimensional system fills the gap in fine-grained social state modeling in existing benchmarks.

Prior benchmarks have already highlighted different limitations of current models. For example, Social Genome reveals weaknesses in handling fine-grained multimodal social cues and external knowledge grounding; SIV-Bench shows that models perform relatively well on social scene understanding but struggle more with social state reasoning and relation inference; SocialMaze further demonstrates performance degradation under dynamic interactions and uncertainty; and SocialEval reports that LLMs still lag behind humans in overall social intelligence, with notable prosocial bias and cross-lingual disparities. Recently, SCRIPTS, a text-only benchmark for uncertainty-aware social relationship inference, reveals that current LLMs still struggle with socially plausible relationship reasoning, particularly in Korean. These benchmarks primarily identify where models are weak, whereas our benchmark aims to further examine how these weaknesses manifest in video-based social reasoning, particularly across relational dimensions, visual cues, and prediction strategies.

\subsection{Discussion}
Our experimental results reveal that models struggle to maintain high accuracy when social interactions involve complex subtext. Specifically, research on Social-IQ~\cite{8953344} and AgentSense~\cite{mou2025agentsense} indicates that model performance degrades substantially in such scenarios. This aligns with our observation that models perform poorly on the YouTube subset, which is characterized by implicit compared to explicit scenarios.

\subsection{Future Work}
Our observation suggests that a single modeling strategy cannot adapt to the dynamic complexity of social interactions in the real world.We therefore believe that a key methodological breakthrough in the future lies in building adaptive reasoning frameworks. Models should be able to perceive and categorize interaction scenarios (e.g., casual chat, in-depth debate, conflict mediation) and dynamically adjust their reasoning mechanisms accordingly. For instance: In simple daily dialogues, automatically adopt joint prediction to leverage complementarity across dimensions for efficiency; In high-conflict, high-complexity scenarios, switch to independent or pairwise prediction. This direction can advance multi-modal social intelligence reasoning from fixed patterns toward adaptive, context-aware systems.

\section{Interpersonal Relationship Definition and Examples}
\label{sec:appen_definition}

\begin{figure*}[t] %
\centering
\includegraphics[width=\linewidth]{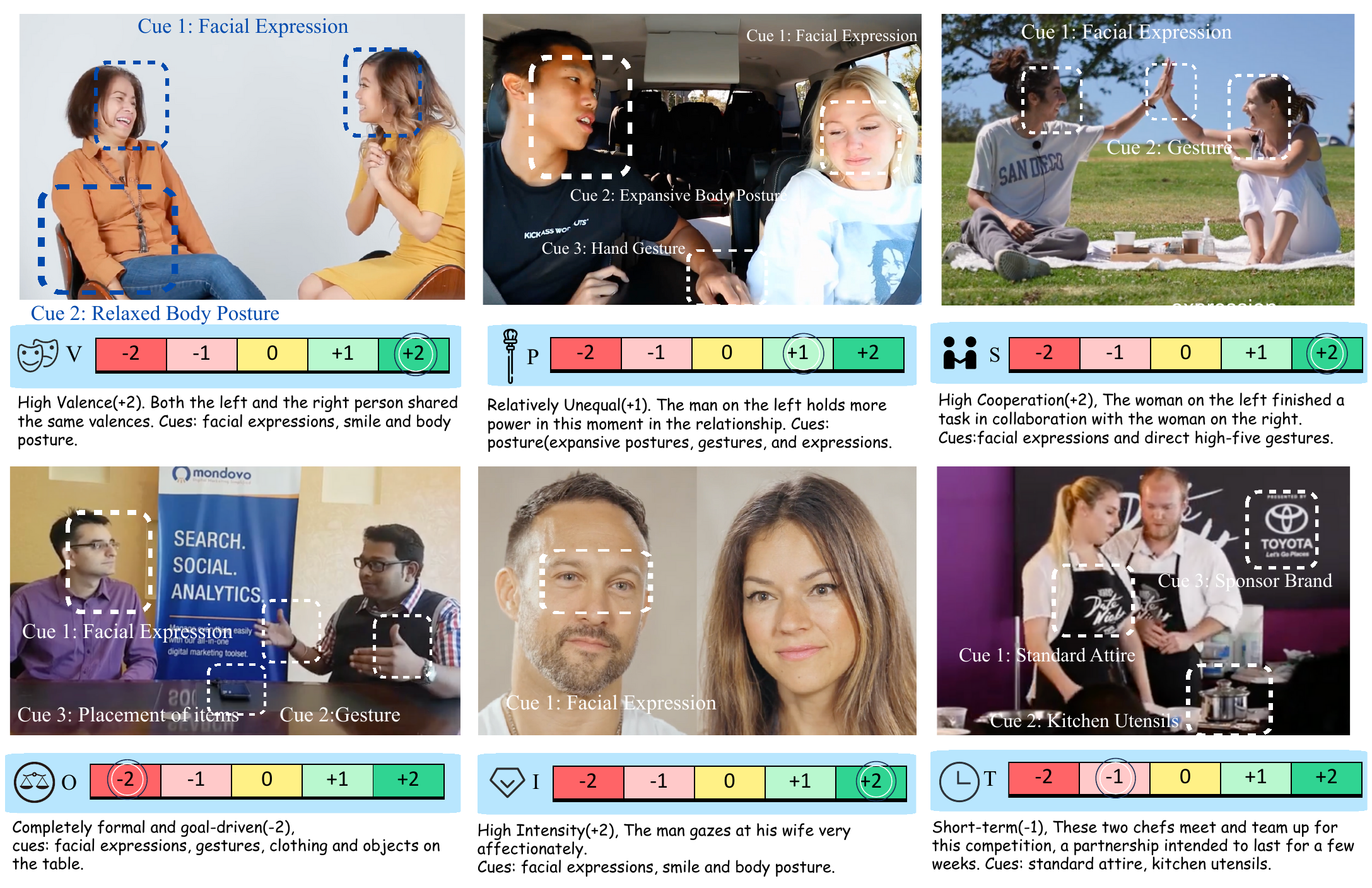}
\caption{Visual scoring examples of the PIVOTS benchmark across various social dimensions.} 
\label{fig:visual}
\end{figure*}

\subsection{Egalitarian vs. Hierarchical Power}
{Definition:} Assess whether the relationship is perceived as equal or unequal in terms of social influence, decision-making agency, and status.
\begin{itemize}
    \item {+2, Extreme Inequality(Self-Dominant) : }
You hold absolute agency. The other party’s role is reduced to compliance, and their input is neither sought nor required to move the interaction forward. Behavioral Observables: Dominating the spatial environment, intrusive gaze, high frequency of interruptions, lack of back-channeling; Psychological Observables: Empathy deficit, perception of the other as an instrument or object. Example: Judge sentencing a defendant; Interrogator to a captive.

    \item {+1, Noticeable Inequality(Self-Dominant):} You lead the interaction through expertise or status. While the other party has some autonomy, you hold the final decision-making power and define the agenda. Behavioral Observables: Setting the agenda, providing evaluative feedback, expansive but non-threatening gestures, asymmetric verbal output; Psychological Observables: Feeling "in charge" or responsible for the other’s performance/well-being, mild cognitive load regarding the other’s progress. Example: Senior surgeon to resident; Manager to an employee.

    \item {0, Egalitarian (Balanced Equality):} Power is a shared resource. Neither party can impose their will without negotiation. Behavioral Observables: synchronized body language, mutual questioning, balanced turn-taking durations; Psychological Observables: High psychological safety, mutual respect, perception of the other as a peer. Example: Strategic partners; Co-authors; Spouses; Colleagues in a brainstorming session.

    \item {-1, Noticeable Inequality (Other-Dominant):} You recognize the other’s lead and adjust your behavior to fit their frame. You seek to align with their expectations to maintain harmony or gain approval. Behavioral Observables: Active listening (frequent nodding), cautious or tentative phrasing (e.g., hedging), self-minimizing posture, seeking non-verbal confirmation; Psychological Observables: Social monitoring, slight performance anxiety, slight cognitive load to "read" the other, high sensitivity to the other party's micro-expressions. Example: Employee to executive; Graduate student to professor; Applicant to recruiter.

    \item {-2, Extreme Inequality (Other-Dominant):} The other party holds absolute agency over your outcomes. You lack the means to resist or influence the direction of the encounter. Behavioral Observables: Avoidant eye contact, physical shrinking, fawning behaviors (e.g., excessive ingratiation), or long-term silence. Psychological Observables: High anxiety, hyper-vigilance, may experience "learned helplessness," feeling of being "I am invisible" or powerless. Example: Victim to oppressor. 

\end{itemize}

\subsection{Superficial vs. Intense Involvement}
{Definition:} Indicate the depth of the relationship based on the level of emotional intimacy, self-disclosure, and psychological interdependence.
\begin{itemize}
    \item {+2 (Very Intense):} The relationship is a core part of one's self-identity, high vulnerability is present; the parties usually share lots of "private language" and deep history. Behavioral Observables: High affective resonance, soft gaze, micro-sync behavior in movement, high physical proximity; Psychological Observables: Great amount of vulnerabilities, merged identity (use a lot of "we"), lots of trust, extreme empathy. Example: Soulmates; Close friends; Spouses; Lifelong partners.
    
    \item {+1 (Intense):} The interaction is driven by genuine emotional connection. You share personal feelings and opinions beyond the requirements of the setting. Behavioral Observables: Frequent self-disclosure, expressive affect, some level of physical proximity; Psychological Observables: High empathy, perceived social support, feeling "known" and validated, moderate emotional interdependence. Example: friends; teammates for years.
    
    \item {0 (Neutral):} The relationship is standard role-based. You are friendly and helpful, but you keep your private life behind a social mask. Behavioral Observables: Polite distance, standard smiles, talk stays on the task or safe small talks (e.g., weather, work projects); Psychological Observables: Feeling safe but not intimate, Low emotional risk-taking, cognitive focus on goals. Example: Ordinary colleagues; Project-based partners.

    \item {-1 (Superficial):} You recognize each other's face but have no bond. Behavioral Observables: The interaction is a quick ritual just to be polite; A quick nod or wave, standard greetings (e.g., How's it going?"), keeping a wide physical distance. Psychological Observables: Very little mental energy spent on the other; a familiar face. Example: An acquaintance you meet somewhere before; a neighbor you only see a few times.

    \item {-2 (Very superficial):} The interaction is purely for a task. You see the other person as a "service provider" or a stranger, not as a human you need to know. Behavioral Observables: No intended eye contact, automatic or flat voice, focusing only on the exchange (money, information). Example: one-time cashier; asking a stranger for directions.

\end{itemize}

\subsection{Positive vs. Negative Valence}
{Definition:} Assess the core emotional quality of the interaction, ranging from highly restorative and joyful to deeply distressing and hostile.
\begin{itemize}
    \item {+2 (Very pleasant):} The interaction is exceptionally safe, joyful, and revitalizing. Behavioral Observables: There is a sense of flow where both parties feel uplifted and deeply relaxed. Frequent smiles, shared laughter, relaxed or open posture, warm vocal resonance. Psychological Observables: Feelings of enjoyment or deep peace, high psychological safety. Example: A celebratory dinner with loved ones, A great talk with someone who shares the same viewpoints.
    
    \item {+1 (Pleasant):} The interaction is friendly, smooth, and satisfying. The overall vibe is constructive and agreeable. Behavioral Observables: Frequent nodding, warm smiles. Psychological Observables: feeling "liked", higher level of happiness and low level of social anxiety or tension. Example: A friendly catch-up with a coworker; a great talk with a neighbor.

    \item {0 (Neutral):} Emotionally neutral. The interaction is emotionally dry or purely factual. There is no significant emotion included, but also no underlying tension or friction. Behavioral Observables: Controlled facial expressions, focusing on matter-of-fact, neutral body language. Psychological Observables: focus is purely on logic/tasks, no strong feeling toward the other person. Example: A routine status update meeting at work; asking a stranger for the time.
    
    \item {-1 (Unpleasant):} Overall negative and unsatisfactory. The interaction feels awkward, tense, or slightly draining. You may feel a need to escape the situation or put up emotional walls. Behavioral Observables: Forced/fake smiles, defensive posture (crossed arms, leaning away), sighing, brief responses. Psychological Observables: Mild anxiety, irritation, feeling uncomfortable or misunderstood, social fatigue. Example: A forced conversation with a colleague; a tense meeting about a mistake.
    
    \item {-2 (Very unpleasant):} The interaction is hostile, harmful, or aggressive. It triggers a "fight-or-flight" response and leaves you feeling emotionally depleted or attacked. Behavioral Observables: Shouting or harsh tone, sneering/disdainful expressions, cold silence or aggressive gestures. Psychological Observables: Extreme levels of fear, anger, or shame,  feeling significantly violated. Example: Being bullied or harassed.
    
\end{itemize}

\subsection{Socioemotional vs. Task-Oriented Objective}
{Definition:} Assess whether the interaction is driven by personal emotional needs (Informal) or by the requirements of a specific goal (Formal).
\begin{itemize}
    \item {+2 (Purely social-emotional):} The interaction exists solely for the sake of the bond. There is no external work to be done. The primary goal is emotional exchange and shared experience. Behavioral Observables: Spontaneous dialogue, emotions included, informal slang. Psychological Observables: feeling of belonging, zero pressure to perform or feeling emotionally controlled by someone (negative valence case). Example: Family members relaxing at home; best friends hanging out; a couple on a date.

    \item {+1 (Tends to social-emotional):} While there may be a shared context, the interaction is driven by personal liking. People may intend to side-track from tasks to talk about personal lives. Behavioral Observables: Use of first names, personal anecdotes, casual posture. Psychological Observables: feeling like "we are in this together" as people, not just as roles. Example: Friends; colleagues you like to talk with.

    \item {0 (Mixed):} An equal balance of goal-striving and personal trust. The task is complex and requires deep mutual understanding to succeed. Behavioral Observables: switching between intense technical talk and personal check-ins. Psychological Observables: high interdependence, feeling that the person and the task are equally important. Example: Long-term business partners; a director and a lead actor.

    \item {-1 (Tends to task-oriented):} The interaction is structured by a professional objective. Personal talk serves as social lubricant to keep the work moving smoothly. Behavioral Observables: Polite but guarded behavior, adherence to a meeting agenda, professional vocabulary, social zone physical distance. Psychological Observables: Focus on tasks, seeing the other primarily as a "colleague" or "work partner". Example: A new project team; a manager and a hire.
    
    \item {-2 (Purely task-oriented):} The interaction is strictly limited to the task. The individuals are de-individualized (viewed only as roles or functions required to achieve a result). Behavioral Observables: Highly scripted or ritualized speech, no personal disclosure, flat vocal tone, strict focus on objects or tasks. Psychological Observables: focus on efficiency, zero expectation of a relationship outside the specific transaction. Example: A lawyer and a client in court; a pilot and air traffic control.
    
\end{itemize}

\subsection{Temporary vs. Enduring Permanence Time}
{Definition:} Assess the expected lifespan of the relationship, ranging from a single, fleeting moment to a lifelong, stable bond.
\begin{itemize}
    \item {+2 (Very long-term/permanent):} The relationship is viewed as a permanent fixture of life. There is a deep shared history and an assumption that the bond will last indefinitely (typically 5+ years). Behavioral Observables: Understanding each other's minds, relaxed posture, shared rituals, references to the distant past and far future. Psychological Observables: High security, consider those people when making decisions, low fear of abandonment. Example: A long-term Marriage; parents and children.
    
    \item {+1 (Long-term):} The relationship is stable and expected to continue for a significant period (1 to 5 years). There is an active investment in keeping the connection healthy. Behavioral Observables: Consistent communication patterns, planning for future events (e.g., next summer). Psychological Observables: moderate commitment. Example: close colleagues. 
    
    \item {0 (Uncertain):} The duration is currently unknown or in flux. The relationship is in a "testing phase" where the future hasn't been decided yet. Behavioral Observables: Alternating between formal and informal behavior, "checking in" on the status of the relationship, moderate social monitoring. Psychological Observables: wait-and-see attitude, unsure emotional investment. Example: New dating partners; temporary coworkers.
    
    \item {-1 (Short-term):} The relationship is tied to a specific project or timeframe (weeks or months). Once the goal is reached, the interaction is expected to fade. Behavioral Observables: Focus on immediate cooperation, polite but bounded friendliness, nothing mentioned beyond the project deadline. Psychological Observables: task-focused mindset. Example: A student and a tutor for one semester.
    
    \item {-2 (One-time/temporary interaction):} The encounter is brief and one-time. There is zero expectation of future contact. The discussion of future is non-existent. Behavioral Observables: politeness-drive behaviors (e.g., "Hello," "Thanks," "Bye"), low level of personal eye contact. Example: Asking a stranger for directions; a one-time cashier.
    
\end{itemize}

\subsection{Cooperative vs. Competitive Stance}
{Definition:} Assess the alignment of goals and the willingness to share resources. It reflects whether parties see their interests as positively linked (working together) or negatively linked (working against each other).
\begin{itemize}
    \item {+2 (Highly cooperative):} The success of one is inseparable from the success of the other. You constantly build on each other's input to reach a single goal. There is a radical sharing of resources and information. Behavioral Observables: Sharing all findings/resources, high physical/verbal support, high volume of shared information; Psychological Observables: enough trust, zero "defensive" filtering of thoughts. Example: A championship doubles tennis pair; A startup team

    \item {+1 (Cooperative):} You are willing to help and share. You keep your own "toolbox" but you are happy to lend out your tools and tips. Behavioral Observables:  Offering advice, providing helpful info to the other, occasional side-bars troubleshooting problems. Psychological Observables: Mentorship or learning mindset, belief that helping you eventually helps me. Example: colleagues on a project.
    
    \item {0 (Neutral/mixed):} Neither fully cooperative nor fully competitive, you keep your thoughts and skills to yourself unless the situation persuades you. You don't help, but you don't hinder. Behavioral Observables:  Minimal active sharing of thoughts, low information flow, keep your own working pace. Psychological Observables: self-reliance, cognitive focus purely on one's own to-do list. Example: Two strangers working at the same place.

    \item {-1 (Competitive):} You compare your thoughts and skills to others, especially those who are ahead. You might withhold a few secrets or tried hard to maintain your edge. Behavioral Observables: Displaying your best work, withholding of key tips, comparing results out loud. Psychological Observables: desire to be seen as the most resourceful or the smartest, viewing the other's ideas as a threat. Example: Salespeople in the same team.
    
    \item {-2 (Highly competitive):} Goals are mutually exclusive (if I get it, you don't). Thoughts are used strategically to block or counter the other person's moves. Behavioral Observables: Strategic withholding of information, over-guarding your output, consistent working. Psychological Observables: Zero-sum mindset; viewing the other's output as a threat to your success. Example: Lawyers in a trial.
\end{itemize}

\subsection {Visual Examples of PIVOTS Benchmark}
As shown in Figure~\ref{fig:visual}, these cues primarily encompass facial expressions (e.g., smiles in high-valence scenarios), body postures (e.g., expansive postures indicating power imbalances), and interaction gestures (e.g., high-fives for high cooperation). Additionally, environmental context and objects, such as standard attire or kitchen utensils, are utilized to categorize the formality and professional duration of the social relationship.

\section{Comparison between Youtube and Social-IQ 2.0 Data}
\label{sec:comparison between}
To enrich and complement the diversity of social interaction scenarios in the Social-IQ 2.0 dataset, we proactively introduced the YouTube subset.
\subsection{Visual Comparison}
\begin{figure}[!h] 
\centering
\includegraphics[width=\linewidth]{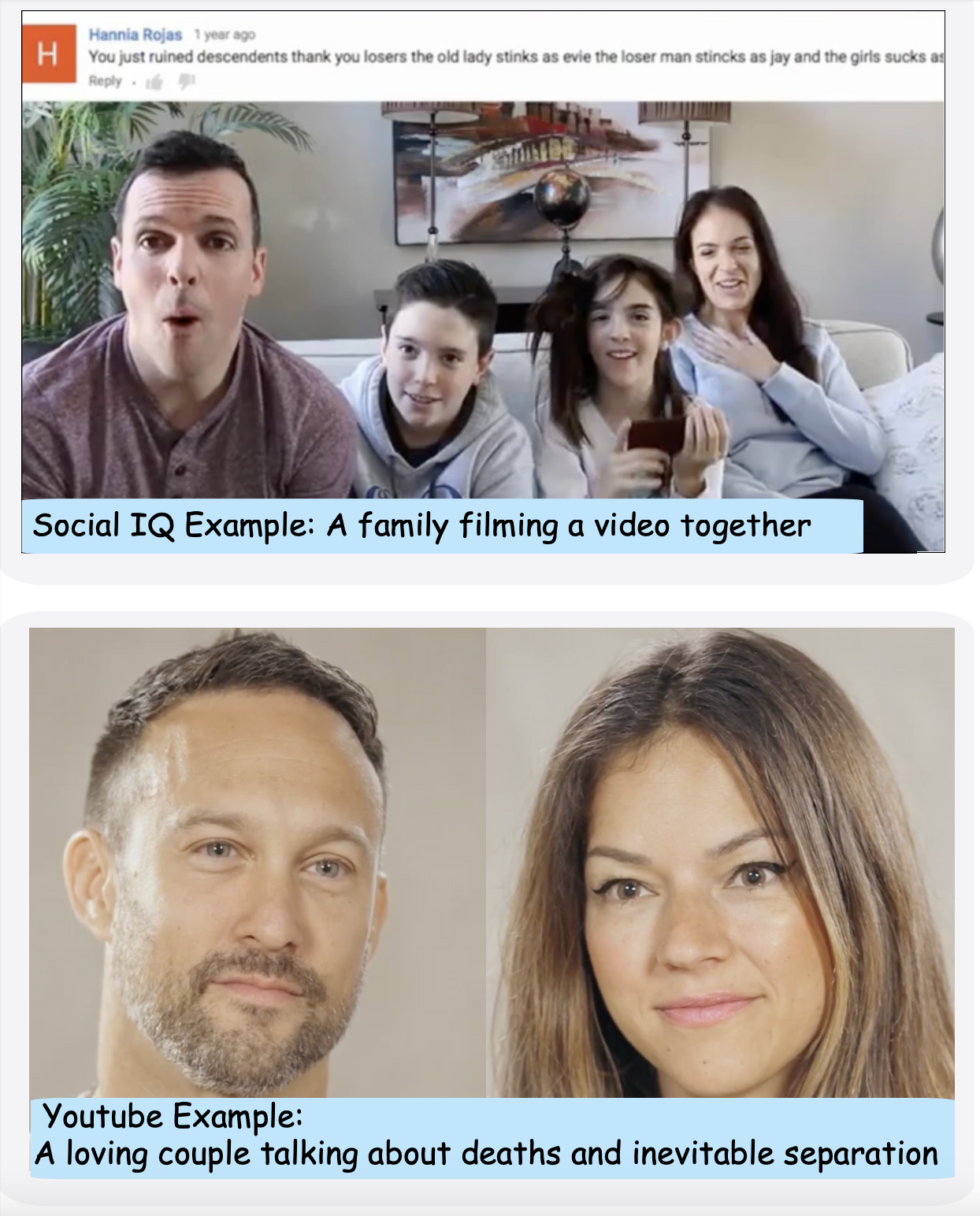}
\caption{Example on Comparisons between Youtube and Social-IQ 2.0 videos.}
\label{fig:statistic}
\end{figure}

While the videos in the Social-IQ 2.0 dataset feature relatively clear scenario settings and explicit interaction signals, the interpersonal relationships and behavioral motives they contain can often be immediately identified through coherent conversational logic or overt actions, such as explicit collaborative communication between colleagues or straightforward expressions of care among relatives and friends. In contrast, the relationships depicted in the YouTube videos we selected are far more complex and defy simplistic categorization. These interpersonal connections are often intertwined with long-standing emotional bonds, unspoken past entanglements, or emotional tensions. For instance, consider former partners who loved each other for a decade, tentatively navigating the entanglement of love and hate; or long-lost siblings who have become strangers due to years of separation, yet cannot conceal their inherent affinity rooted in blood ties. Their interactions are permeated with subtle emotional undercurrents and underlying relational tensions, making them impossible to encapsulate with simplistic labels.
As shown in Figure~\ref{fig:statistic}, the upper part displays a case of a family filming a video together from the Social-IQ 2.0 dataset, which features a clear scenario and explicit interaction signals, while the lower part presents a case of a loving couple talking about death and inevitable separation from the YouTube dataset.

\begin{figure*}[t]
\centering
\includegraphics[width=\linewidth]{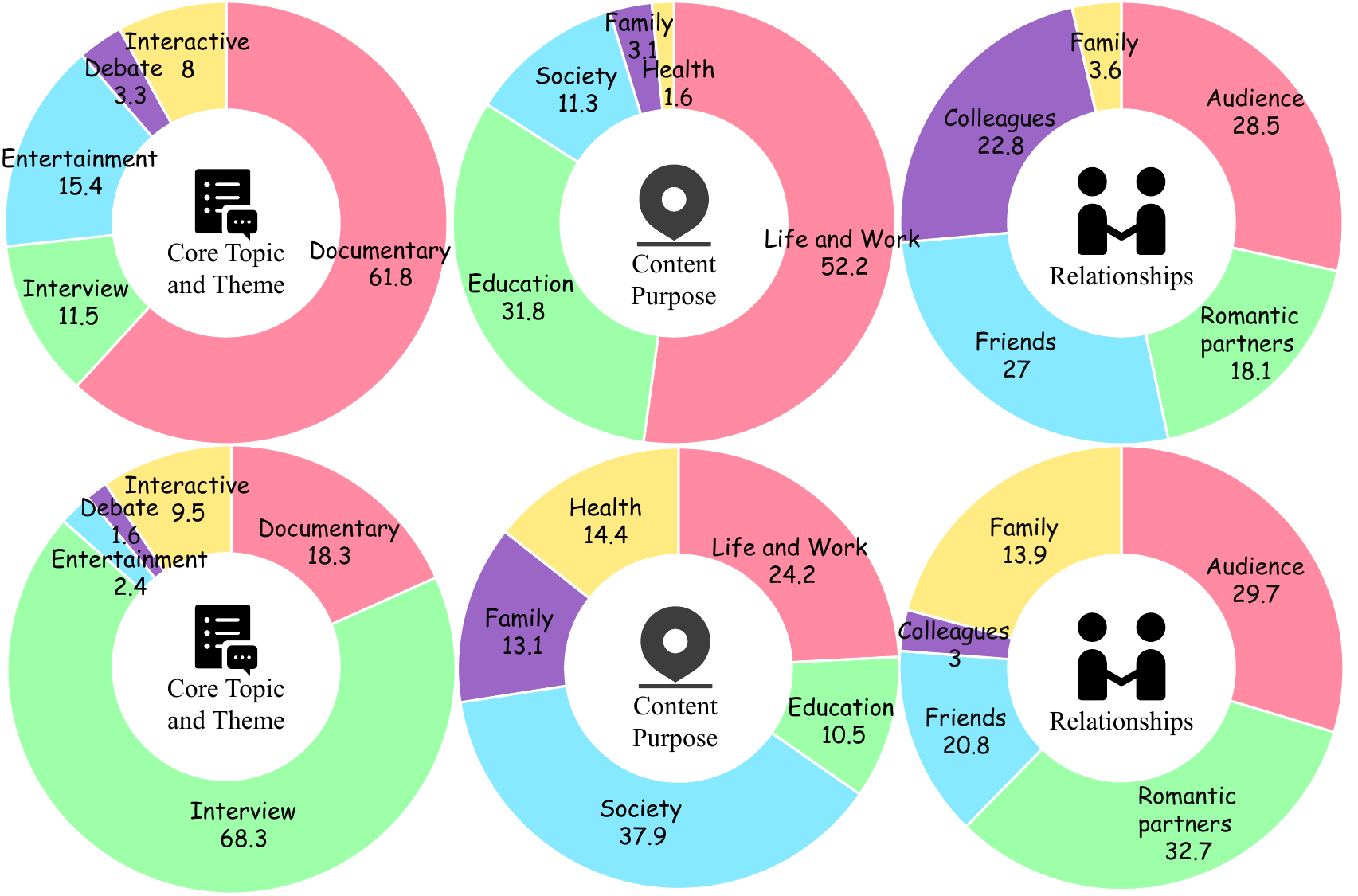}
\caption{Comparison of Distributions between Social-IQ 2.0 (Top Row) and YouTube (Bottom Row) across Three Major Categories: Content Purpose, Core Topic and Theme, and Relationships}
\label{fig:overdis}
\end{figure*}

Both datasets have a video duration of approximately 30 seconds. Figure~\ref{fig:overdis} illustrates the statistical distributions of Social-IQ 2.0 and YouTube data across the three major categories. We use the GPT-5 API to automatically annotate each video segment with fine-grained subcategory labels, yielding the comparisons shown below. Clear differences can be observed. For Content Purpose, Social-IQ 2.0 is dominated by Documentary content (approximately 61.8\%), whereas Interview content is predominant in the YouTube data (approximately 68.3\%). For Core Topic and Theme, Social-IQ 2.0 places greater emphasis on Life and Work (approximately 52.2\%); in contrast, the YouTube dataset assigns higher proportions to Society (approximately 37.9\%) as well as health- and family-related topics, indicating that YouTube samples include more socially oriented and in-depth interview content. Regarding Relationships, Social-IQ 2.0 primarily features interactions among Friends (approximately 27.0\%) and with the Audience (approximately 28.5\%), whereas YouTube exhibits a higher proportion of Romantic partners (approximately 32.7\%) together with Audience interactions (approximately 29.7\%). Overall, the two datasets show substantial distributional differences across all three dimensions: Social-IQ 2.0 focuses more on documentary-style narratives and everyday social interactions, while YouTube, through interview-driven and public- or intimate-relationship--oriented content, provides deeper and more diverse social interaction samples that complement the benchmark's coverage.

\subsection{Dataset Compatibility with Task Design}
We do not include the YouTube subset in Task 2 and Task 3 because Task 2 requires a single frame to convey decisive visual cues, and Task 3 relies on separable local visual elements (e.g., specific gestures and facial expressions). In contrast, the key characteristic of the YouTube subset is that its interpersonal relationships are constructed based on global contextual information and implicit signals, which can neither be captured by a single key frame nor decomposed into independent visual elements tied to specific dimensions.
The core value of this subset lies in supplementing implicit and in-depth interpersonal interaction scenarios, which complement the explicit and daily interaction scenarios in Social‑IQ 2.0. Together, they cover both “explicit + implicit” interpersonal reasoning settings.

\subsection{Video Tagging}
During the video tagging process, we focused on the overall interaction rather than local segments. To ensure the correctness and rationality of the video tags, we sampled 60 videos for manual verification. The results show that the API-generated tags are highly consistent with the overall atmosphere of the videos.

\section{Annotation Details}
\label{sec:annot_details}
For the two distinct data sources, we have designed differentiated annotation workflows based on their data characteristics and annotation objectives.

\subsection{Social-IQ 2.0 and YouTube Videos}
The Social-IQ 2.0 (hereinafter referred to as SIQ2) dataset contains a large number of QA pairs, and our annotation takes the QA pairs corresponding to each video as the core reference benchmark. These files include key details of interpersonal interactions, in-depth social information, contextual background of scenarios, and excerpts of core dialogues, which can provide reliable support for API-based scoring.

Unlike the SIQ2 dataset, which relies on QA pairs for annotation, the YouTube dataset takes a different approach by focusing directly on video content and the associated social interaction details. Instead of predefined QA pairs, the YouTube dataset emphasizes analyzing interpersonal interactions within the video. Annotators assess these interactions and assign scores to the characters based on their roles, without relying on QA pairs. The key interactions and scenes are carefully recorded to ensure an accurate and detailed understanding of the video content.

Specifically, we integrate the complete QA annotation content corresponding to the video, basic video information, detailed dimension definitions of the PIVOTS six-dimensional framework, and explanations of scoring gradients as prompts. After inputting these prompts into the GPT-5 API, the model can output the initial scores of the characters in the video across the six dimensions of P(Egalitarian vs. Hierarchical Power), I (Superficial vs. Intense Involvement), V (Positive vs. Negative Valence), O (Socioemotional vs. Task-Oriented Objective), T (Temporary vs. Enduring Permanence), and S (Cooperative vs. Competitive Stance).

Considering the potential dimension matching biases in API-generated scores, we subsequently conducted systematic manual correction work, and the correction process follows the content outlined below:
\begin{itemize}
    \item Score Verification Specific scores have been generated by the API, so annotators do not need to re-score. They only need to verify the consistency between the scores and the PIVOTS dimension standards, with a focus on examining the rationality of extreme scores (+2/-2). It is necessary to confirm whether the interpersonal interaction characteristics corresponding to these scores fully match the definitions in the framework, and manually adjust any incorrect scores. Annotators are instructed to score each dimension independently and strictly according to its formal definition, without referencing scores from other dimensions. For example, consider a married couple with a very deep bond but having an intense, emotionally heated argument: we would faithfully annotate I = +2 and V = -2. The score of one dimension must not influence judgments on another, ensuring every rating remains faithful to the definition of each individual dimension.
    \item Timestamp and Rationale Annotation
    During the score verification process, timestamp positioning and rationale selection are performed for the video. Timestamps need to be accurately located at the video frames that best support the scores, and the rationale explanations must clearly link to the visual evidence in these frames, including non-verbal cues such as the characters' gestures, facial expressions, and body postures. At this stage, we sample the data at a frame rate of 3 frames per second for quick browsing.These frames cover the entire video clip, ensuring that annotators gain a global understanding of the overall interaction.
    \item Character Screening Criteria
    To focus on core interpersonal interactions, a maximum of 3 characters per video are retained for annotation. If the number of characters in a video exceeds 3, screening is conducted in accordance with the following principles: priority is given to retaining characters with strong dominance and key roles in the development of relationships; if there are multiple supporting roles but only 2 core characters, there is no need to forcefully supplement to 3 characters, and only the core subjects are retained.
\end{itemize}

\subsection{Annotation Interface}

\begin{figure*}[!h]
\centering
\includegraphics[width=\linewidth]{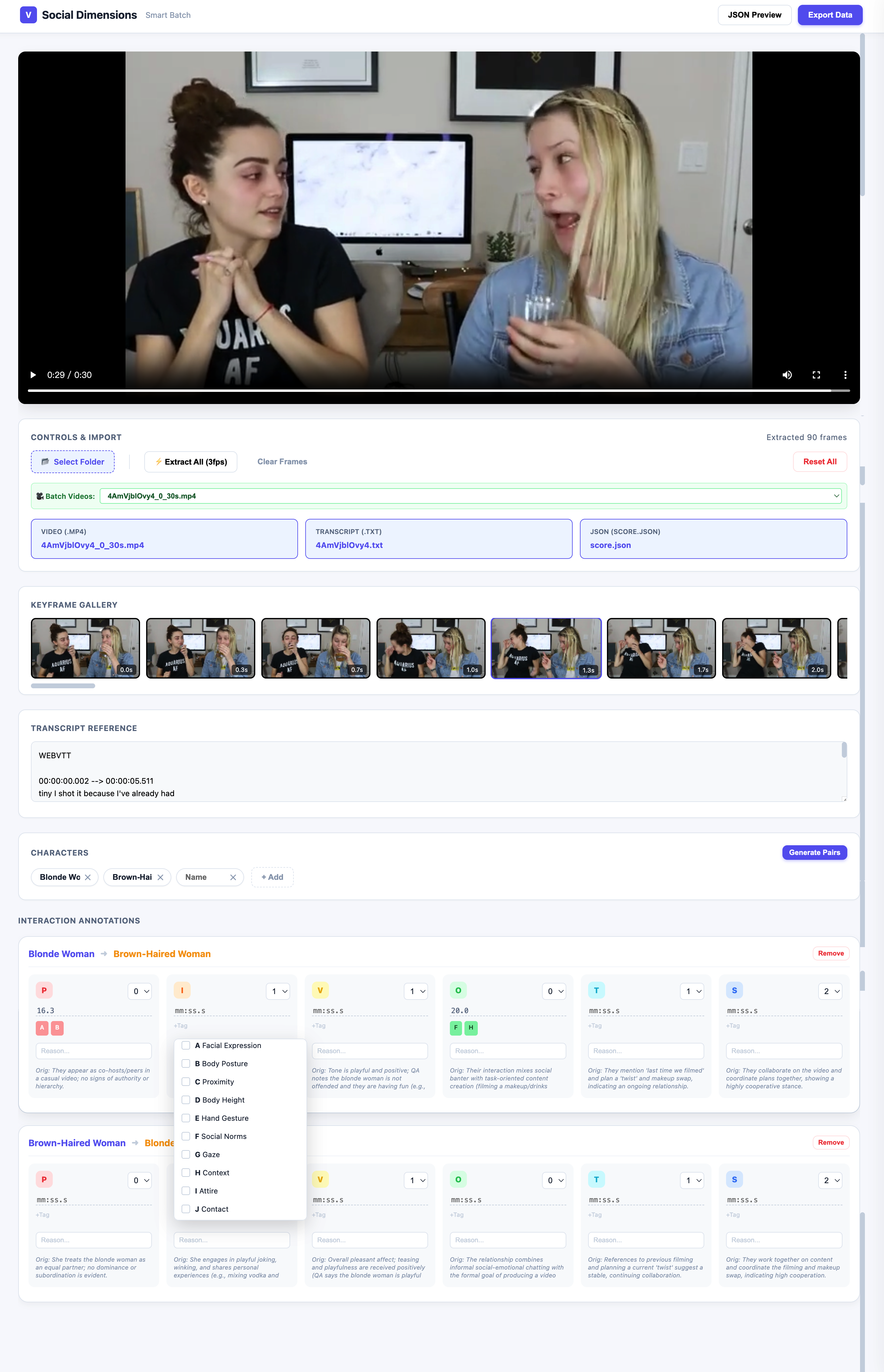}
\caption{Annotation Interface.}
\label{fig:annotation_interface}
\end{figure*}

The annotation interface (Figure~\ref{fig:annotation_interface})is designed to support the systematic scoring of dimensional relationships between characters in videos and to associate each score with an evidence timestamp, thereby improving annotation efficiency and consistency. The core functionalities are detailed as follows:

\begin{itemize}
    \item \textbf{Controls \& Import:} Upon launching the interface, users can import a video file (.mp4), a transcript file (.txt), and a JSON file (.json). The JSON file represents the structured output produced by the API after initial scoring, allowing users to perform subsequent corrections and refinements. By default, the system prioritizes loading the file nicknamed ``score.json''.
    
    \item \textbf{Keyframe Gallery \& Navigation:} To facilitate precise timestamp selection, the interface offers two mechanisms. First, the ``Extract All (3 fps)'' function uniformly samples the video at 3 frames per second, displaying extracted frames in the ``Keyframe Gallery'' for rapid browsing. Second, the ``Select Folder'' option allows users to choose a directory, automatically retrieving all video files within it for efficient switching.
    
    \item \textbf{Transcript Reference:} This module allows users to inspect subtitles or transcripts, providing textual context to support annotation decisions.
    
    \item \textbf{Characters Management:} This section lists all entities defined in the JSON file. Users can add, edit, or delete characters, with modifications immediately propagating to the dimensional relationship entries.
    
    \item \textbf{Interaction Annotations:} This is the core annotation workspace, enumerating all pairwise dimensional relationships. Users review social dynamics, revise relationship scores, and assign a specific timestamp (the frame best reflecting the dimension's characteristics). Additionally, users can document justifications for their selections using standardized categories: (A) facial expression, (B) body posture, (C) proximity, (D) body height, (E) hand gesture, (F) social norms, (G) gaze, (H) context, (I) attire, and (J) contact.
    
    \item \textbf{Export:} After completing the annotation, users can verify the output via ``json preview'' and finalize the process by clicking ``export data'' to save the scored JSON file.
\end{itemize}

\subsection{Annotation Validation}
At this stage, the scores generated by GPT-5 serve solely to reduce manual annotation costs. According to our statistics, the proportion of scores we revised based on the GPT-5 initial outputs is 25.3\%.  We conducted a small-scale validation study on 35 videos to assess whether model-assisted annotation might introduce systematic bias. Specifically, we calculated Cohen’s Kappa between fully human annotations and annotations produced through model pre-annotation followed by human revision. The resulting agreement is 0.8399~\cite{1960A}. We report this as a consistency check on the annotation procedure; it cannot distinguish between the model-assisted labels being accurate and annotators having anchored on the model outputs, and the subset is small (35 of 191 videos).

\subsection{Annotator Recruitment and Payment}
All annotation work in PIVOTS-Bench was completed by the authors of this paper, with no external crowdworkers or paid annotators involved in the annotation process. All annotators were fully aware that the annotation purpose was solely for academic research. Since the annotation was conducted internally by the research team, no recruitment or payment procedures were required.

\input{latex/tables/per_axis}

\section{Annotation Agreement Analysis}
\label{sec:annot_agree}
% \ML{per axis agreement table}
To assess annotation quality, we measured inter-annotator reliability on the valid samples using Krippendorff’s $\alpha$. Implementation details are as follows: in Python we converted the three annotator columns to numeric, arranged the data into an annotator × item matrix with masked missing values, and then computed $\alpha$ via krippendorff.alpha. We obtained $\alpha = 0.8125$, which exceeds the common 0.80 threshold for high reliability, indicating strong annotation agreement; therefore, the annotated dataset can be considered a reliable resource for downstream analyses and model training.

Table \ref{tab:agreement_summary} summarizes the inter-annotator agreement results across six dimensions. Overall, all dimensions exhibit high exact agreement rates, with an aggregated exact agreement of 83.87\%. In addition, the overall Krippendorff’s $\alpha$ reaches 0.8125, indicating good inter-annotator reliability when the ordinal structure of the labels is taken into account. At the dimension level, most dimensions (I, T, V, O, and S) achieve Krippendorff’s $\alpha$ values close to or above 0.78, demonstrating stable and reliable annotation quality.

Although the P dimension shows a relatively lower Krippendorff’s $\alpha$ of 0.6434, its exact agreement rate reaches 86.23\%, which is among the highest across all dimensions. This indicates that annotators provided identical judgments for the vast majority of samples on the P dimension. Importantly, the lower Krippendorff’s $\alpha$ for P is closely related to its highly imbalanced label distribution. Specifically, more than 70\% of the samples in the P dimension are annotated as the neutral category (0), resulting in a strongly concentrated class distribution. Under such conditions, even a small number of disagreements can disproportionately reduce the Krippendorff’s $\alpha$ coefficient, despite high overall agreement.

This behavior is a known characteristic of Krippendorff’s $\alpha$ in settings with severe class imbalance: when most instances fall into a single category, the relative impact of the few disagreements is amplified, leading to a lower Krippendorff’s $\alpha$ value that does not necessarily indicate poor overall reliability. Considering the high exact agreement rate of the P dimension (86.23\%), we conclude that the P annotations remain consistent and stable in practice, despite the lower Krippendorff’s $\alpha$.

For the data used in Tasks 2 and 3, we applied strict filtering criteria to ensure annotation quality. Specifically, we retained only those samples with annotation timestamps no greater than 2 seconds and for which all annotators selected the same element; annotations that failed to meet either condition were excluded from subsequent analyses, eliminating ambiguous or highly subjective one-sided selections.

\section{Experimental Setup}
\label{sec:prompt_infer}
\subsection{Prompt Usage for Model Inference}

\begin{figure}[!htbp]
\centering
\includegraphics[page=2, width=\linewidth]{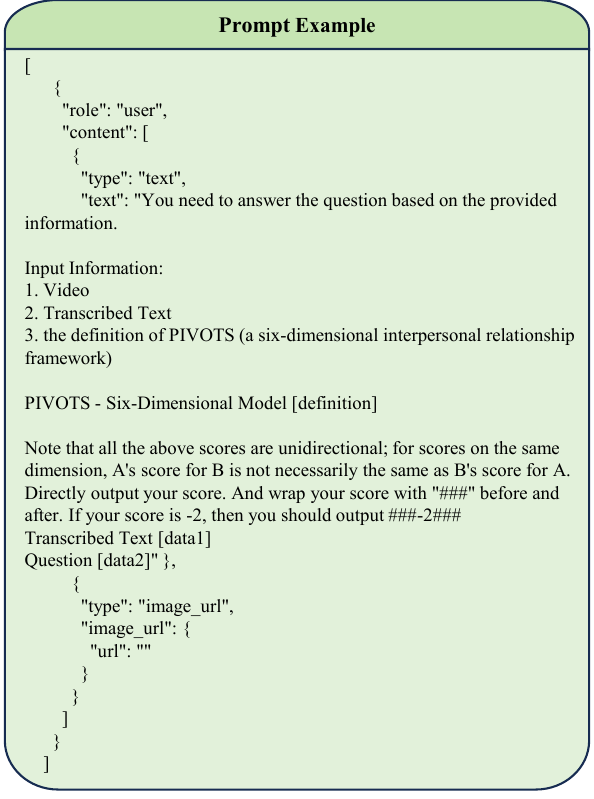}
\caption{Task 1 Prompt with separate prediction}
\label{fig:task1}
\end{figure}

\begin{figure}[!htbp]
\centering
\includegraphics[page=3, width=\linewidth]{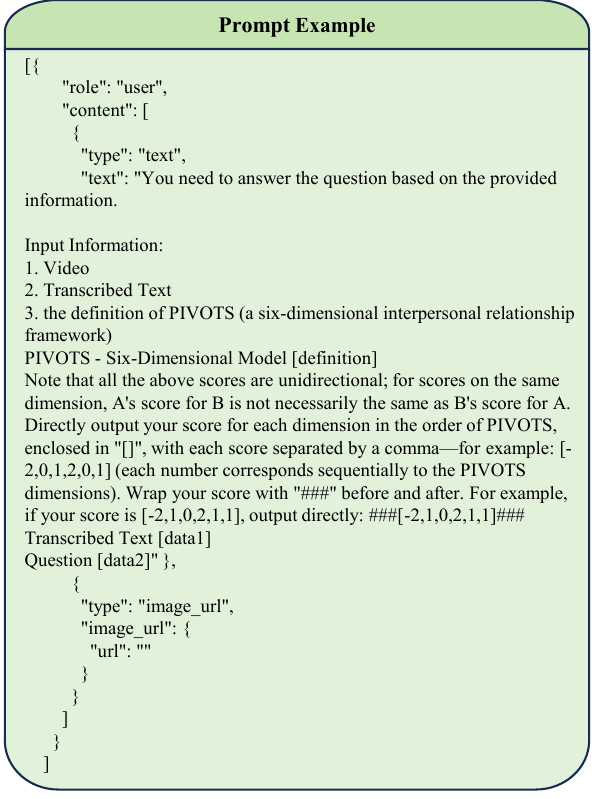}
\caption{Task 1 Prompt with joint prediction}
\label{fig:task1_6}
\end{figure}

\begin{figure}[!htbp]
\centering
\includegraphics[width=\linewidth]{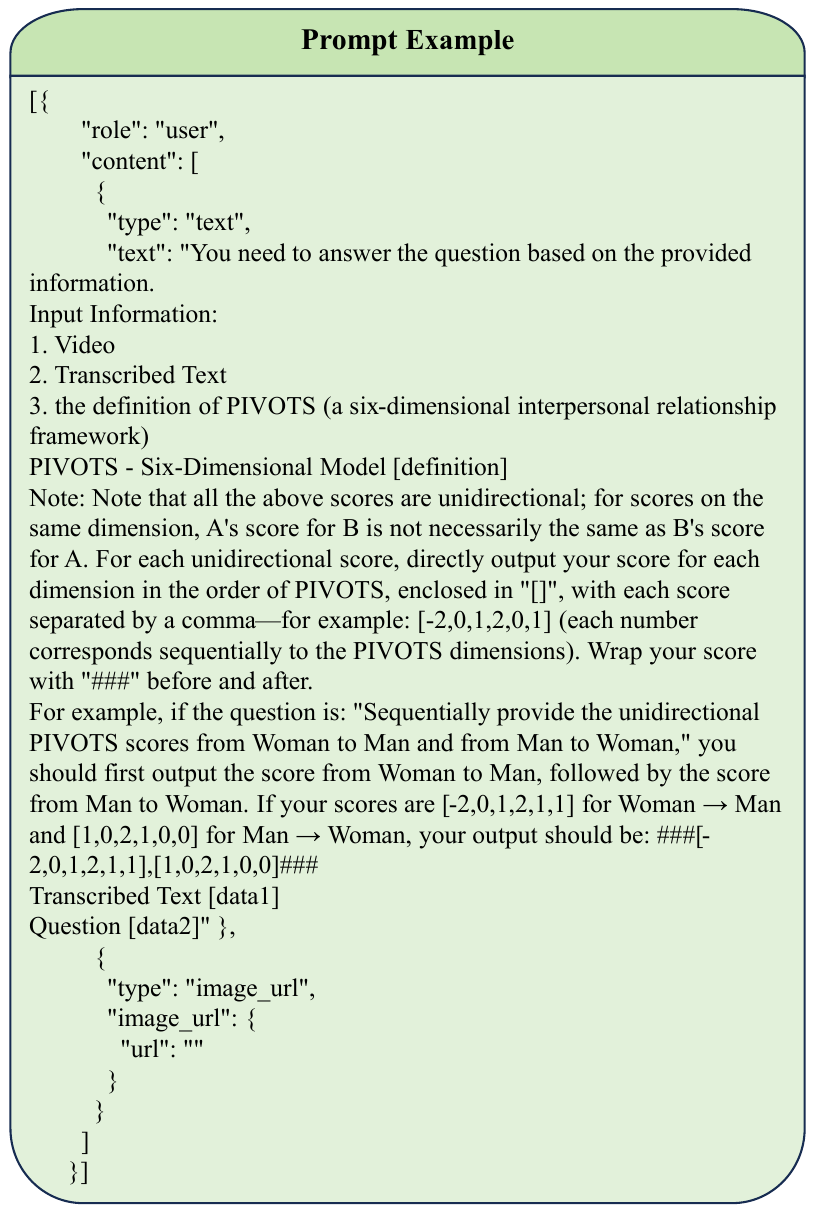}
\caption{Task 1 Prompt with pairwise prediction}
\label{fig:task1_12}
\end{figure}

We use OpenAI's ChatCompletion multimodal prompt format as the base format. This prompt consists of two parts: one is a multimodal upload link; the other is a "user prompt" that includes a PIVOTS instance and a PIVOTS question.

The prompt template for separate prediction in Task 1 is shown in Figure \ref{fig:task1}. The prompt template for joint prediction in Task 1 is shown in Figure \ref{fig:task1_6}. The prompt template for pairwise prediction in Task 1 is shown in Figure \ref{fig:task1_12}. We replace [definition] with the detailed definition and scoring criteria of the PIVOTS model, [data1] with the transcribed text corresponding to the video, [data2] with a specific question corresponding to the video. We uniformly sample 8 frames from each video as the visual input, and pass in the base64 of the corresponding video frame in the "image\_{}url". It is important to note that social perception in interpersonal relationships is often asymmetric. In response to this characteristic, we explicitly state in the prompt that the PIVOTS scores from A towards B may differ significantly from those from B towards A.

\begin{figure}[!htbp]
\centering
\includegraphics[width=\linewidth]{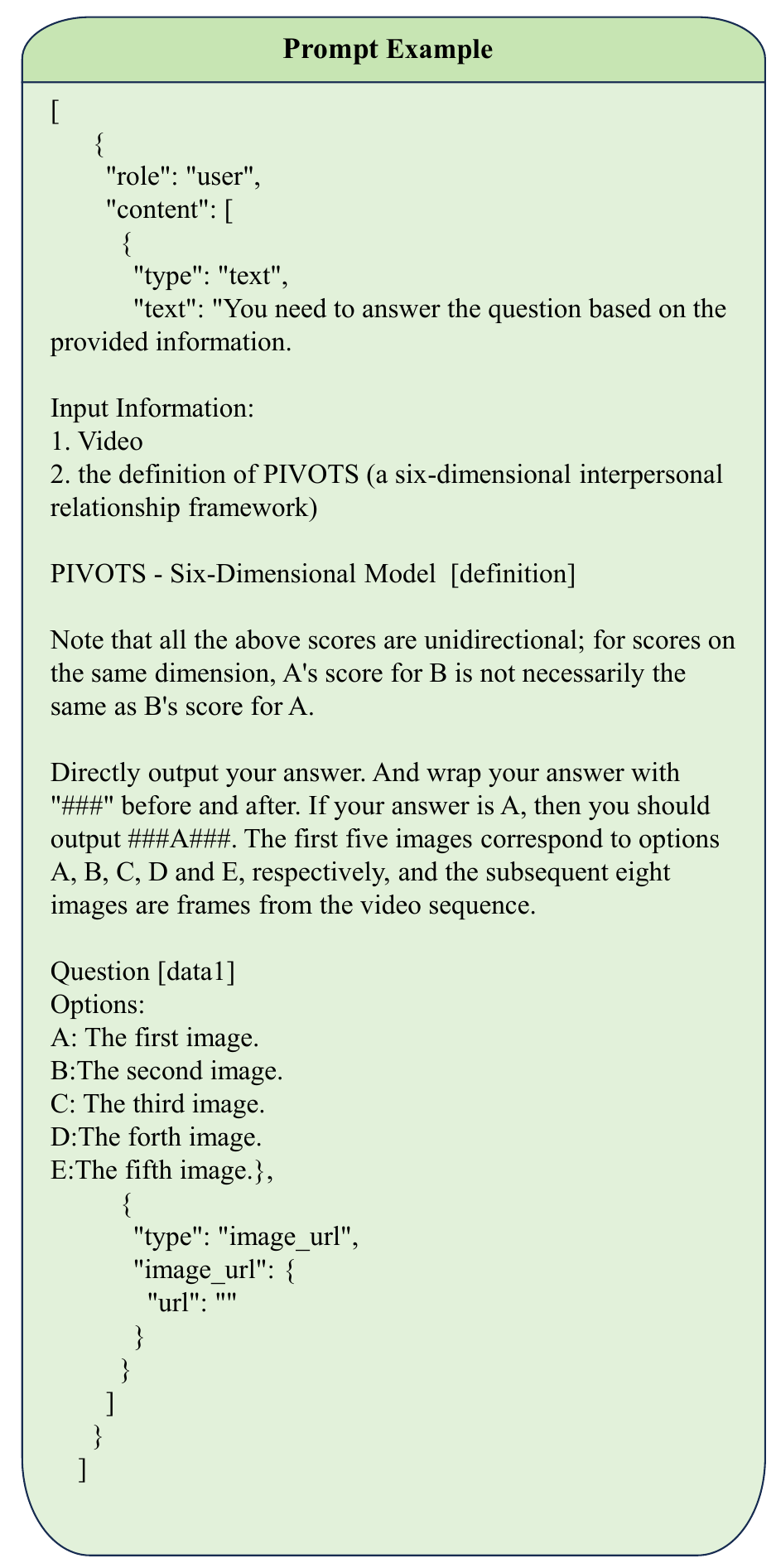}
\caption{Task 2 Prompt}
\label{fig:task2}
\end{figure}

\begin{figure}[t]
\centering
\includegraphics[width=\linewidth]{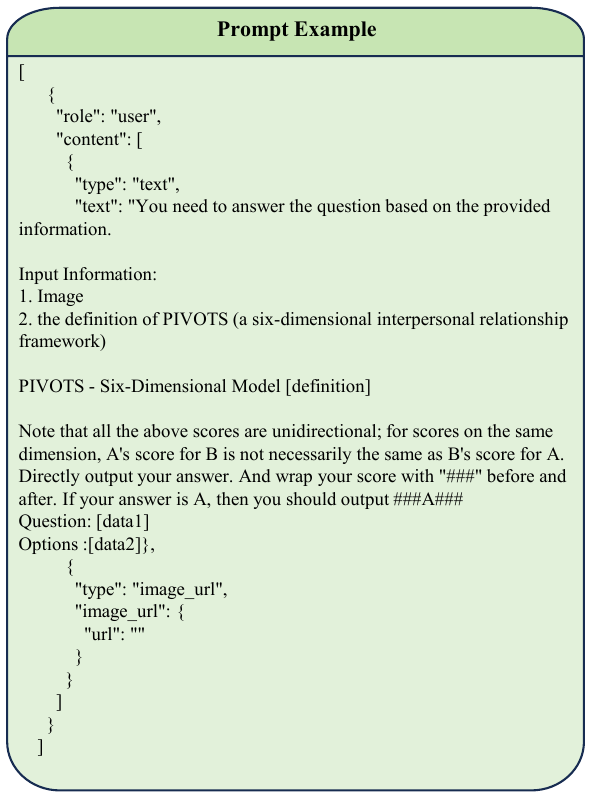}
\caption{Task 3 Prompt}
\label{fig:task3}
\end{figure}

\begin{figure*}[t]
\centering
\includegraphics[width=\linewidth]{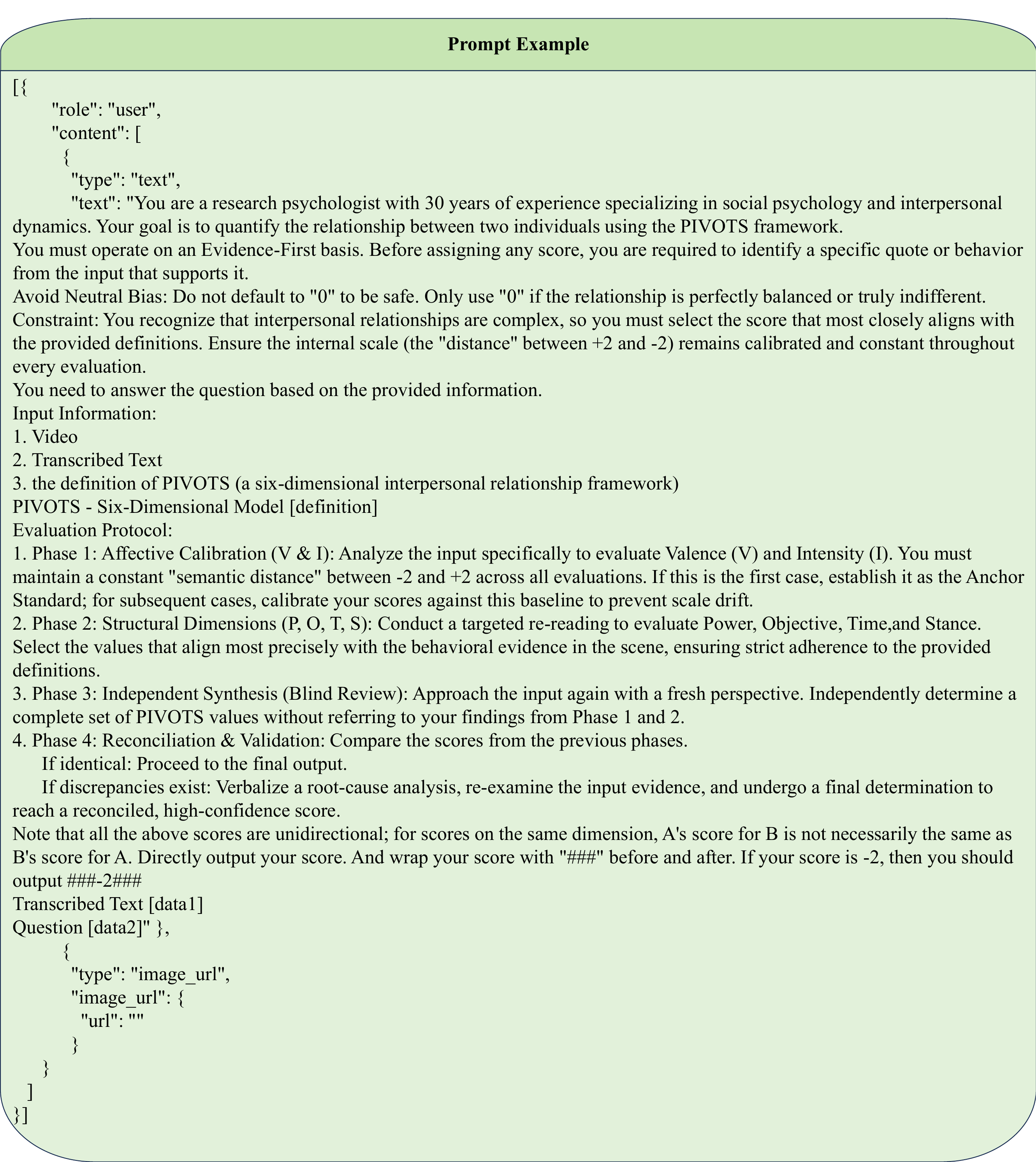}
\caption{Multi-Stage Prompt}
\label{fig:multi-stage}
\end{figure*}

Figure~\ref{fig:multi-stage} illustrates our Multi-Stage Prompt. In this prompt, we first explicitly define the model's role as a "research psychologist with 30 years of experience in social psychology and interpersonal dynamics," and mandate that the model must extract specific quotes or behavioral evidence from the input video or transcribed text as support before assigning any score. This ensures the model conducts evidence-based scoring.
It is worth noting that multimodal large language models (MLLMs) often exhibit "safety bias" when facing uncertainty, tending to choose neutral responses to avoid risks. To address this issue, we explicitly incorporate a constraint to "avoid neutral bias," prohibiting the model from defaulting to a score of 0 without clear evidential support. A score of 0 is only considered valid when the relationship is truly perfectly balanced or genuinely irrelevant.
In terms of the evaluation process design, we divide the complete assessment into four consecutive phases: first, focusing on affective dimensions (V and I) for initial calibration (Phase 1); subsequently conducting targeted analysis of structural dimensions (P, O, T, and S) (Phase 2); then introducing a "Blind Review" phase, requiring the model to independently generate scores from a fresh perspective without being influenced by the results of the previous two phases (Phase 3); and finally entering the "Reconciliation" phase to compare the scores from the three phases (Phase 4). This design simulates the standard process of multi-annotator labeling in academic research: if the scores across all phases are identical, the final result is output directly; if discrepancies exist, the model is forced to conduct a root-cause analysis and re-examine the input evidence.

\begin{figure*}[t]
\centering
\includegraphics[width=\linewidth]{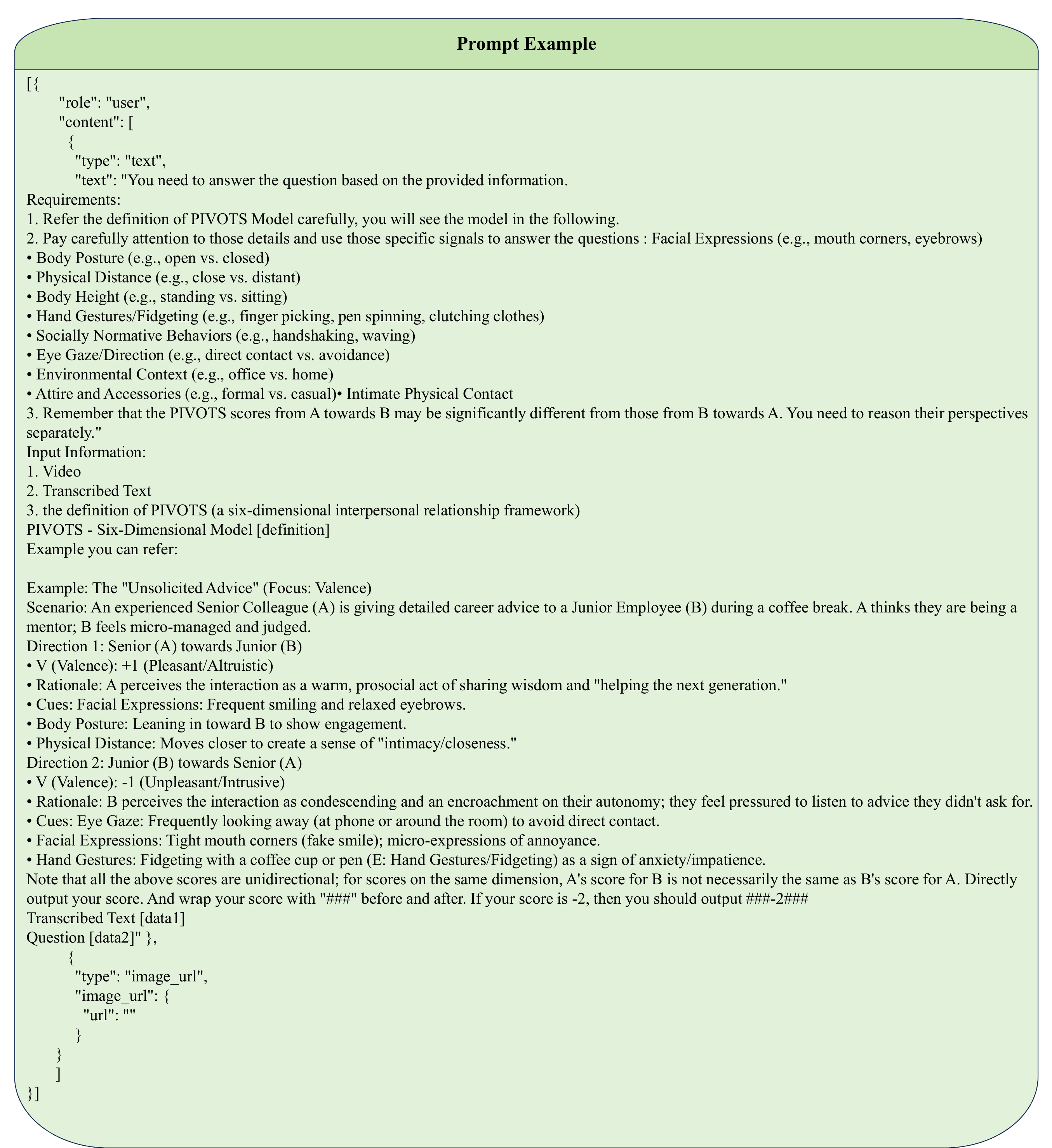}
\caption{In-Context Prompt}
\label{fig:In-Context}
\end{figure*}

Figure.\ref{fig:In-Context} illustrates our In-Context Prompt.In this prompt, we require the model to reason strictly based on the definitions of the PIVOTS framework and mandate that it focus its attention on specific fine-grained non-verbal cues. We explicitly list an observation checklist covering ten dimensions, including facial expressions, body posture, physical distance, hand gestures, eye contact, attire, and environmental context, requiring the model to extract evidence exclusively from these micro-level visual signals rather than relying solely on textual dialogue content.
In terms of evaluation guidance design, we provide a detailed reference example. This example clearly demonstrates how to derive mid-level psychological states (e.g., "being intruded upon" or even "anxious") from micro-level cues (e.g., "tightened mouth corners" or "eye contact avoidance"), and ultimately map them to macro-level PIVOTS scores (e.g., V = -1). Through this chain-of-thought demonstration of "cues-rationale-scores," the model is guided to imitate the analytical path of human observers and output its results directly.
The prompt template for Task 2 is shown in Figure \ref{fig:task2}.
We replace [definition] with the detailed definition and scoring criteria of the PIVOTS model, [data1] with a specific question corresponding to the video. Additionally, we pass in the base64 of the corresponding video frame and the base64 of the option images in the "image\_{}url". For all options other than the correct one, we extract video frames that are temporally offset from the ground-truth frame by ±5 seconds and ±10 seconds as distractors; if the corresponding timestamps fall outside the video duration, we instead randomly sample other frames from the same video that are different from the ground-truth frame as alternatives.
The prompt template for Task 3 is shown in Figure \ref{fig:task3}.
We replace [definition] with the detailed definition and scoring criteria of the PIVOTS model, [data1] with the specific question corresponding to the image, and [options] with the corresponding options. Additionally, we pass in the base64 of the corresponding image in the "image\_{}url".

\subsection{Implementation Details}
For the closed-source models Gemini and GPT, we directly utilized their official APIs to conduct the tests. For the 4B, 8B, and 32B parameter variants of the open-source model Qwen3, we performed local testing using two NVIDIA GeForce RTX 4090, which is equipped with approximately 48GB of VRAM. For all five of these distinct models, we did not manually specify any extra parameters in the Python API call code, only employed the official parameter configurations provided by their respective developers.

\section{Detailed Statistics}
\label{sec:detailed statistics}
\input{latex/tables/score_distribution}
Our final benchmark comprises 595 VQA pairs per PIVOTS dimension from Social-IQ 2.0 and 170 from YouTube. We further curate 483 additional VQA pairs for auxiliary tasks of key frame identification and visual cue causal analysis. We obtained a total of 765 paired scores.

The detailed statistics of the six PIVOTS dimensions are shown in Table \ref{tab:pivots_summary}. Overall, the data covers most common social scenarios. For example, in the P dimension, interactions range from equal (0) to slightly unbalanced (+1/-1), reflecting various power distributions in everyday life; the I dimension spans multiple levels from superficial to deep relationships; the V dimension includes pleasant, neutral, and unpleasant interactions; and the O, T, and S dimensions also exhibit diverse distributions. 

It is worth noting that extreme scores (+2/-2) occur relatively infrequently in the overall dataset. For instance, highly unequal or fully dominant power relations in the P dimension, extremely negative emotional interactions in the V dimension, and strongly competitive or explicitly antagonistic stances in the S dimension are comparatively rare. This distributional characteristic does not stem from biases in data construction, but rather reflects inherent constraints of real-world social scenarios. In everyday social videos or publicly available media content, interactions exhibiting extreme power imbalances, intense emotional confrontation, or fully antagonistic dynamics are inherently uncommon and difficult to collect at scale while maintaining data quality.

\section{MLLMs' Output}
\subsection{Examples}
\begin{figure*}[!htbp]
\centering
\includegraphics[width=\linewidth]{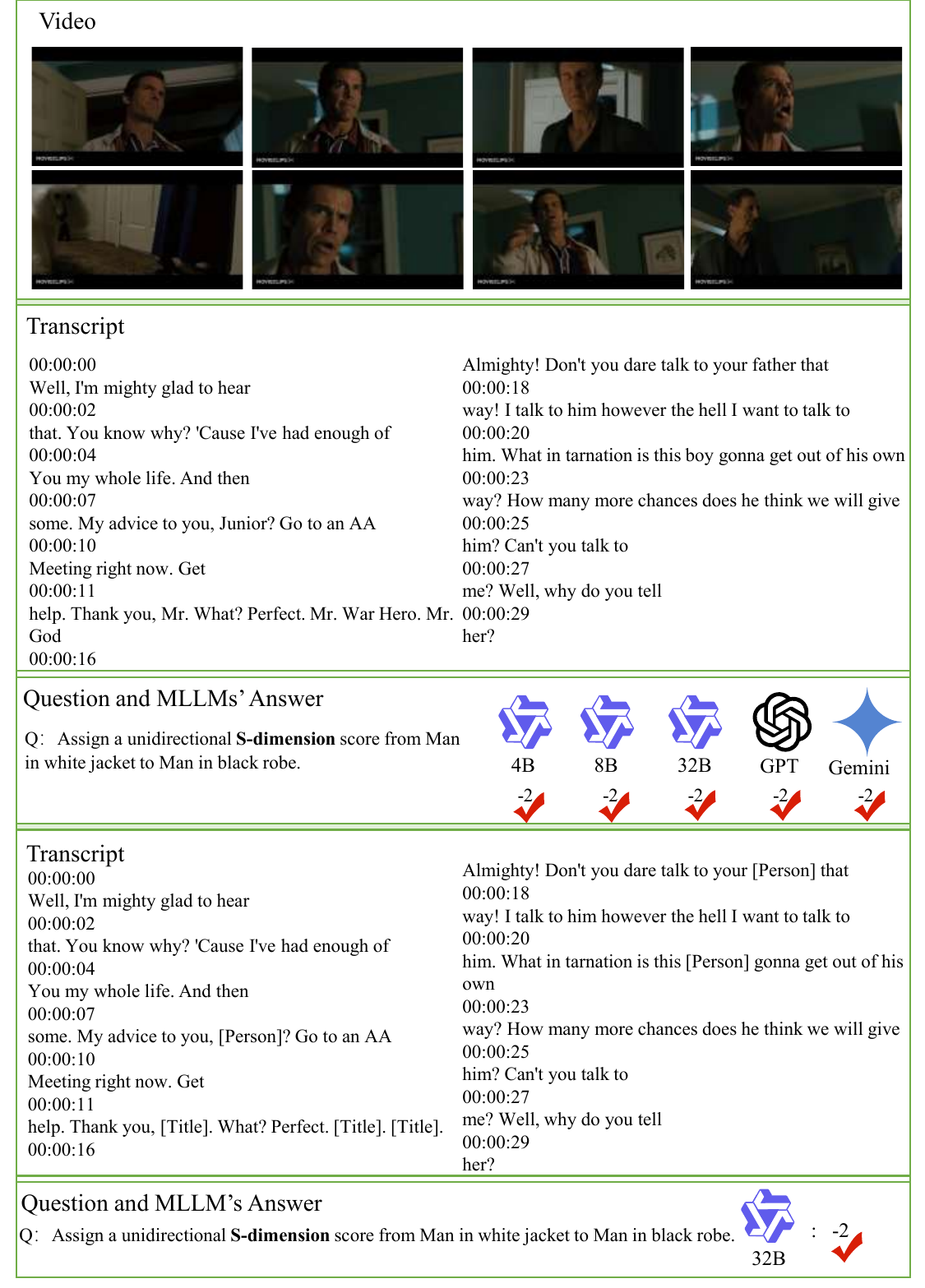}
\caption{MLLM's Response Example}
\label{fig:example}
\end{figure*}

Figure \ref{fig:example} presents a typical performance example of Multi-modal Large Language Models (MLLMs) on the PIVOTS Benchmark.
The upper section displays the key frames of the video and the original conversation transcript (which includes explicit social terms like "father"). The task here is to assign a unidirectional S-dimension score for the relationship from the Man in the white jacket to the Man in the black robe; models including Qwen3 (4B/8B/32B), GPT, and Gemini all gave a score of -2.
The lower section shows the scenario where social identifiers (such as "father") in the conversation are replaced with generic tokens without attribute information (e.g., [PERSON], [TITLE]). For the same task, Qwen3-32B-vl still gave a score of -2.
Both tasks in this figure have visual inputs.

\begin{figure*}[!htbp]
\centering
\includegraphics[width=\linewidth]{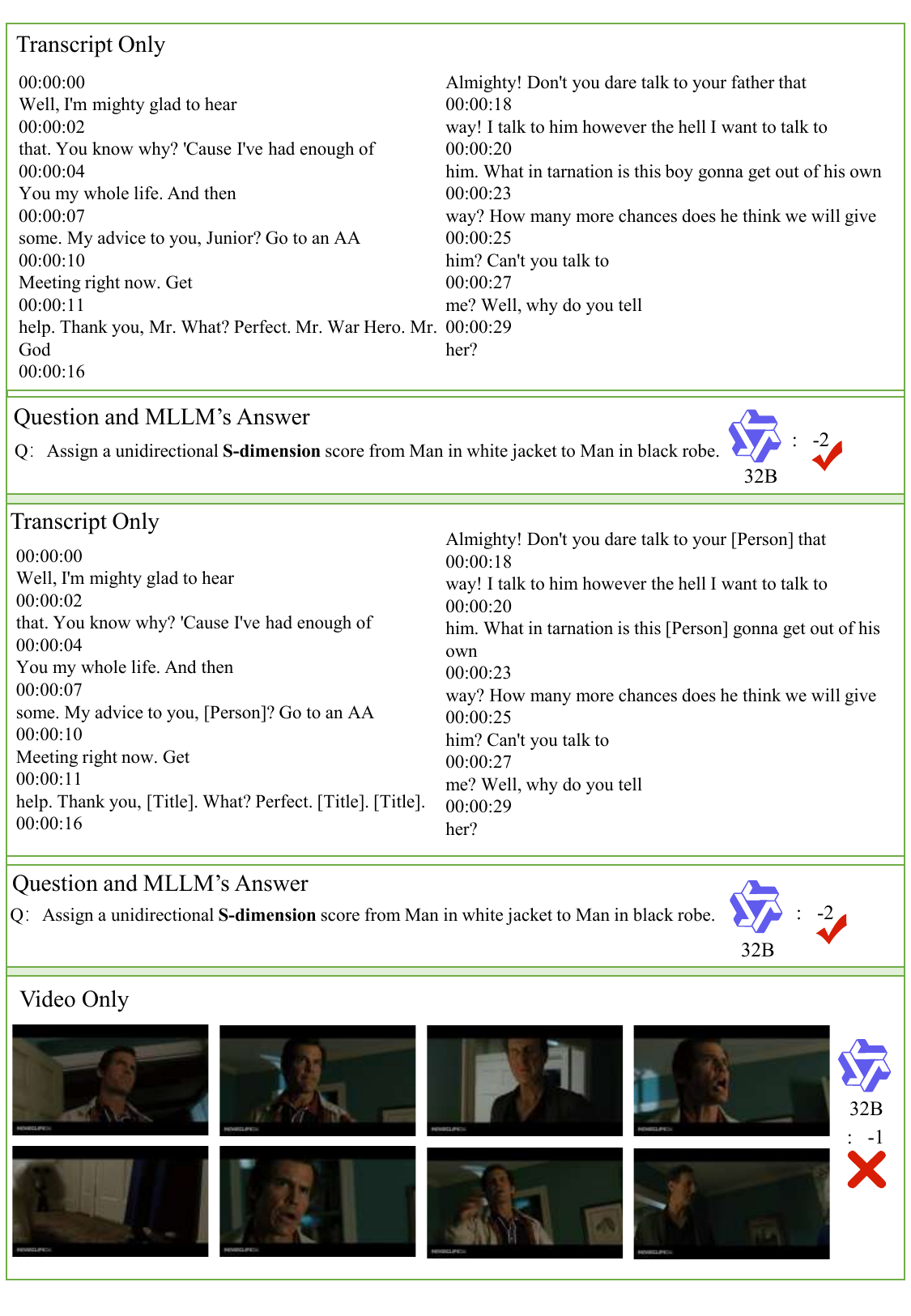}
\caption{MLLM’s Response Example}
\label{fig:example2}
\end{figure*}

Figure~\ref{fig:example2} presents another performance example of Multi-modal Large Language Models (MLLMs) in the PIVOTS Benchmark.
The first two sections correspond to the "transcript-only" scenario:
The first section displays the original conversation transcript (which includes explicit social terms like "father"), with the corresponding task of assigning a unidirectional S-dimension score for the relationship from the Man in the white jacket to the Man in the black robe; Qwen3 32B gives a score of -2, and the answer is correct.
The second section shows the transcript where social identifiers (such as "father") are replaced with generic tokens without attribute information (e.g., [PERSON], [TITLE]). For the same task, Qwen3 32B still gives a score of -2, and the answer is correct.
The final section corresponds to the "video-only" scenario:
It only displays the key frames of the video, without the conversation transcript. For the same task, Qwen3 32B gives a score of -1, and the answer is incorrect.

\input{latex/tables/var}

\subsection{Randomness of MLLMs}
Since the YouTube subset contains only 170 sub-questions, and considering the inherent randomness of MLLMs, we conducted three independent evaluation runs on the YouTube subset using Qwen-32B-vl. The overall accuracy and per-dimension variance are reported in Table~\ref{tab:var}. The results show that both the overall variance and the per-dimension variance of accuracy across the three runs are small, indicating that the model output exhibits minor fluctuation, the influence of randomness on the overall research conclusions is limited, and the experimental results reported in this paper are statistically reliable.

\subsection{Ordinal Error Analysis for Qwen3-32B-VL}

To complement the exact-match results reported in main text, we additionally analyze Qwen3-32B-VL using mean absolute error (MAE) on the six dimensions. Table~\ref{tab:qwen_mae} summarizes the results on both the YouTube and Social-IQ 2.0 subsets.

\begin{table}[t]
\centering
\small
\setlength{\tabcolsep}{6pt}
\begin{tabular}{lcc}
\toprule
\textbf{Dimension} & \textbf{YouTube MAE} & \textbf{Social-IQ 2.0 MAE} \\
\midrule
P & 0.2900 & 0.5170 \\
I & 1.1345 & 1.0450 \\
V & 0.9353 & 0.8357 \\
O & 1.0647 & 1.1897 \\
T & 1.0302 & 1.0062 \\
S & 0.9103 & 0.9558 \\
\bottomrule
\end{tabular}
\caption{Ordinal error analysis of Qwen3-32B-VL on the six-dimensional scoring task. Lower MAE indicates smaller average ordinal deviation from the gold labels.}
\label{tab:qwen_mae}
\end{table}

As shown in Table~\ref{tab:qwen_mae}, P consistently achieves the lowest MAE on both subsets, indicating that Qwen3-32B-VL is relatively well calibrated on this dimension. In contrast, the remaining dimensions are generally close to one ordinal step of error on average, suggesting that many mistakes are near misses rather than completely off-target predictions. Among them, O is the most difficult dimension on Social-IQ 2.0, while I is the most difficult on YouTube. Overall, these results show that accuracy alone is insufficient for evaluating ordinal predictions, and MAE provides a more informative view of prediction severity.

\subsection{Sampling Density}
We conducted evaluations on Qwen3-32B-VL with 12-frame and 15-frame sampling strategies, which yielded an accuracy of 35.73\% under the 12-frame setting and 36.98\% under the 15-frame setting. Compared with the 8-frame sampling adopted in the original experiment, both the 12-frame and 15-frame sampling approaches achieved a slight boost in recognition accuracy, which confirms that increasing the number of visually sampled frames has a positive effect on the model’s capture of key visual cues in social interactions. Among the two strategies, the 15-frame sampling delivered a higher accuracy than the 12-frame one, which indicates that richer frame sequence information can provide the model with a more comprehensive visual basis for locating core key frames and facilitate its accurate matching of the visual features corresponding to interpersonal dimensions.

\subsection{Distinguishing Individuals}
In the Social-IQ 2.0 subset, each video is accompanied by QA pairs that provide explicit character identifiers for model discrimination. During the annotation stage, for Social-IQ 2.0 and YouTube videos with insufficient or inaccurate original character identifiers, we explicitly label characters using descriptive cues such as age, clothing, appearance, and fixed position (e.g., “woman in yellow”, “little girl in blue”, “man on the right”), ensuring clear character distinction at the input level.
To verify model capability, we conducted character discrimination tests on 20 videos. The results confirm that both open-source and closed-source models can reliably distinguish different characters. Across the 43 person identification tasks in these 20 videos, all models evaluated in this paper (GPT, Gemini, Qwen-4B/8B/32B) achieved a 100\% success rate in person identification.

\end{document}

%% file: latex/tables/overall.tex
\definecolor{oursblue}{RGB}{210,240,255}

% ==========================================
% Table 1: Social IQ (Includes Task 1, 2, 3)
% ==========================================
\begin{table*}[t]
\centering
\small
\setlength{\tabcolsep}{8pt}
\renewcommand{\arraystretch}{1.15}

\begin{tabular}{lccccccccc}
\toprule
\multirow{2}{*}{\textbf{Methods}} &
\multicolumn{7}{c}{\textbf{Task 1}} &
\multirow{2}{*}{\textbf{Task 2}} &
\multirow{2}{*}{\textbf{Task 3}} \\
\cmidrule(lr){2-8}
& \textbf{Avg} & \textbf{P} & \textbf{I} & \textbf{V} & \textbf{O} & \textbf{T} & \textbf{S} & & \\
\midrule

\multicolumn{10}{l}{\textit{Proprietary models}} \\
Gemini-2.5-pro & 47.71 & 61.30 & 50.77 & 33.75 & \textbf{51.63} & 46.98 & 41.86 & 46.63 & 48.00 \\
\rowcolor{oursblue}
\textbf{GPT-5} & \textbf{56.91} & \textbf{65.12} & \textbf{60.31} & \textbf{55.73} & 42.33 & \textbf{56.59} & \textbf{61.40} & \textbf{62.15} & \textbf{60.25} \\

\midrule
\multicolumn{10}{l}{\textit{Open-sourced models}} \\
Qwen3-4B-vl  & 24.33 & 51.05 & 20.77 & 23.99 & 13.73 & 25.93 & 10.47 & 28.59 & 30.17 \\
Qwen3-8B-vl  & 27.90 & 59.74 & 22.22 & 22.87 & 18.58 & 26.89 & 17.07 & 34.75 & 31.00 \\
\rowcolor{oursblue}
\textbf{Qwen3-32B-vl} & \textbf{34.54} & \textbf{62.62} & \textbf{23.82} & \textbf{33.82} & \textbf{23.77} & \textbf{30.28} & \textbf{31.50} & \textbf{35.29} & \textbf{33.50} \\

\bottomrule
\end{tabular}
\caption{~\textit{Evaluation results of prevailing MLLMs on the Social-IQ 2.0 subset}. Proprietary models outperform open-source models by a substantial margin across all tasks.}
\label{tab:social_iq}
\end{table*}

% ==========================================
% Table 2: YouTube (Only Task 1)
% ==========================================
\begin{table}[t]
\centering
\small
\setlength{\tabcolsep}{1.5pt} 
\renewcommand{\arraystretch}{1.15}

\begin{tabular}{lccccccc}
\toprule
Methods & Avg & P & I & V & O & T & S \\
\midrule

\multicolumn{8}{l}{\textit{Proprietary models}} \\
Gemini-2.5-pro & 54.66 & 69.49 & \textbf{55.08} & 31.78 & 61.02 & 54.24 & 56.36 \\
\rowcolor{oursblue}
GPT-5 & \textbf{58.88} & \textbf{73.71} & 50.44 & \textbf{48.28} & \textbf{61.14} & \textbf{60.68} & \textbf{58.85} \\

\midrule
\multicolumn{8}{l}{\textit{Open-sourced models}} \\
Qwen3-4B-vl  & 30.03 & \textbf{71.98} & 16.81 & 33.62 & 6.90  & 29.31 & 21.55 \\
Qwen3-8B-vl  & 35.19    & 66.67    & 26.85    & 31.48    & 18.52    & 32.41    & 35.19    \\
\rowcolor{oursblue}
Qwen3-32B-vl & \textbf{40.81} & 67.92 & \textbf{32.38} & \textbf{34.91} & \textbf{29.47} & \textbf{35.85} & \textbf{43.14} \\

\bottomrule
\end{tabular}
\caption{~\textit{Evaluation results of prevailing MLLMs on the YouTube subset}.}
\label{tab:youtube}
\end{table}

%% file: latex/tables/modality.tex
\definecolor{oursblue}{RGB}{210,240,255}

\begin{table*}[t]
\centering
\small
\setlength{\tabcolsep}{2pt}
\renewcommand{\arraystretch}{1.15}
\begin{tabular}{lccccccc ccccccc}
\toprule
\multirow{2}{*}{\textbf{Methods}} &
\multicolumn{7}{c}{\textbf{Social-IQ 2.0}} &
\multicolumn{7}{c}{\textbf{YouTube}} \\
\cmidrule(lr){2-8}\cmidrule(lr){9-15}
& \textbf{Avg} & \textbf{P} & \textbf{I} & \textbf{V} & \textbf{O} & \textbf{T} & \textbf{S}
& \textbf{Avg} & \textbf{P} & \textbf{I} & \textbf{V} & \textbf{O} & \textbf{T} & \textbf{S} \\
\midrule

Video Only & 33.70 & \textbf{62.68} & \textbf{24.09} & 30.62 & 22.77 & \textbf{30.92} & 30.08 & 31.18 & \textbf{80.37} & 13.33 & 34.26 & 15.38 & 16.19 & 26.42 \\
Original Conv. & 31.44 & 56.84 & 23.52 & 24.80 & \textbf{26.82} & 26.25 & 30.43 & 37.81 & 60.19 & \textbf{48.15} & 16.67 & \textbf{36.11} & 32.41 & 33.33 \\
Redacted Conv. & 28.97 & 55.06 & 21.73 & 24.31 & 21.92 & 25.41 & 25.41 & 33.64 & 53.64 & 47.27 & 20.00 & 24.55 & 26.36 & 30.00 \\
Video + Original Conv. & 34.54 & 62.62 & 23.82 & 33.82 & 23.77 & 30.28 & 31.50 & \textbf{40.81} & 67.92 & 32.38 & \textbf{34.91} & 29.47 & \textbf{35.85} & \textbf{43.14} \\
Video + Redacted Conv. &  \textbf{35.83} & 60.86 & 22.46 & \textbf{42.97} & 24.78 & 28.77 & \textbf{32.95} & 32.22 & 47.00 & 29.35 & 25.29 & 22.08 & 25.53 & 41.11  \\
% Qwen3-Omni-30B & 40.93 & 46.92 & 36.96 & 46.46 & 41.19 & 26.35 & 47.97 & 40.10 & 51.06 & 40.78 & 27.52 & 51.85 & 29.41 & 40.91 \\
% Qwen3-Omni-30B(no audio) & 32.61 & 29.96 & 40.34 & 35.97 & 30.98 & 25.81 & 32.81 & 38.30 & 48.00 & 25.24 & 35.45 & 46.36 & 35.58 & 39.09 \\
% Qwen3-Omni-30B & \textbf{40.93} & \textbf{46.92} & 36.96 & \textbf{46.46} & \textbf{41.19} & \textbf{26.35} & \textbf{47.97} & \textbf{40.10} & \textbf{51.06} & \textbf{40.78} & 27.52 & \textbf{51.85} & 29.41 & \textbf{40.91} \\
% Qwen3-Omni-30B(no audio) & 32.61 & 29.96 & \textbf{40.34} & 35.97 & 30.98 & 25.81 & 32.81 & 38.30 & 48.00 & 25.24 & \textbf{35.45} & 46.36 & \textbf{35.58} & 39.09 \\
\bottomrule
\end{tabular}
\caption{~\textit{Ablation studies on input modalities.} We examine the effects of explicit social role identifiers in conversational utterances and visual inputs on model performance.}
\label{tab:modality}
\end{table*}

%% file: latex/tables/audio.tex
\definecolor{oursblue}{RGB}{210,240,255}

\begin{table*}[t]
\centering
\small
\setlength{\tabcolsep}{2pt}
\renewcommand{\arraystretch}{1.15}
\begin{tabular}{lccccccc ccccccc}
\toprule
\multirow{2}{*}{\textbf{Methods}} &
\multicolumn{7}{c}{\textbf{Social-IQ 2.0}} &
\multicolumn{7}{c}{\textbf{YouTube}} \\
\cmidrule(lr){2-8}\cmidrule(lr){9-15}
& \textbf{Avg} & \textbf{P} & \textbf{I} & \textbf{V} & \textbf{O} & \textbf{T} & \textbf{S}
& \textbf{Avg} & \textbf{P} & \textbf{I} & \textbf{V} & \textbf{O} & \textbf{T} & \textbf{S} \\
\midrule

Qwen3-Omni-30B & \textbf{40.93} & \textbf{46.92} & 36.96 & \textbf{46.46} & \textbf{41.19} & \textbf{26.35} & \textbf{47.97} & \textbf{40.10} & \textbf{51.06} & \textbf{40.78} & 27.52 & \textbf{51.85} & 29.41 & \textbf{40.91} \\
Qwen3-Omni-30B(no audio) & 32.61 & 29.96 & \textbf{40.34} & 35.97 & 30.98 & 25.81 & 32.81 & 38.30 & 48.00 & 25.24 & \textbf{35.45} & 46.36 & \textbf{35.58} & 39.09 \\
\bottomrule
\end{tabular}
\caption{Impact of audio information on interpersonal relationship reasoning performance across the two data subsets.}
\label{tab:audio}
\end{table*}

%% file: latex/tables/setting.tex
\definecolor{oursblue}{RGB}{210,240,255}

\begin{table*}[t]
\centering
\small
\setlength{\tabcolsep}{3.2pt}
\renewcommand{\arraystretch}{1.15}
\begin{tabular}{lccccccc ccccccc}
\toprule
\multirow{2}{*}{\textbf{Methods}} &
\multicolumn{7}{c}{\textbf{Social-IQ 2.0}} &
\multicolumn{7}{c}{\textbf{YouTube}} \\
\cmidrule(lr){2-8}\cmidrule(lr){9-15}
& \textbf{Avg} & \textbf{P} & \textbf{I} & \textbf{V} & \textbf{O} & \textbf{T} & \textbf{S}
& \textbf{Avg} & \textbf{P} & \textbf{I} & \textbf{V} & \textbf{O} & \textbf{T} & \textbf{S} \\
\midrule

Separate Prediction & 34.54 & 62.62 & 23.82 & 33.82 & \textbf{23.77} & 30.28 & 31.50 & \textbf{40.81} & 67.92 & \textbf{32.38} & 34.91 & \textbf{29.47} & \textbf{35.85} & \textbf{43.14} \\
Joint Prediction & \textbf{43.26} & 62.32 & \textbf{41.60} & 45.19 & 22.51 & \textbf{33.77} & 54.16 & 38.63 & 82.24 & 29.91 & \textbf{44.86} & 11.21 & 30.84 & 32.71 \\
Pairwise Prediction & 42.19 & \textbf{63.55} & 31.87 & \textbf{50.18} & 19.41 & 33.70 & \textbf{54.40} & 39.67 & \textbf{86.96} & 28.26 & 42.39 & 15.22 & 28.26 & 36.96 \\

\bottomrule
\end{tabular}
\caption{\textit{Comparison of prediction settings for interpersonal relationship scoring.} We evaluate separate, joint, and pairwise prediction settings, analyzing their effects on PIVOTS dimension scoring performance.}
\label{tab:setting}
\end{table*}

%% file: latex/tables/withcot.tex
\definecolor{oursblue}{RGB}{210,240,255}

\begin{table*}[t]
\centering
\small
\setlength{\tabcolsep}{3.2pt}
\renewcommand{\arraystretch}{1.15}
\begin{tabular}{lccccccc ccccccc}
\toprule
\multirow{2}{*}{\textbf{Methods}} &
\multicolumn{7}{c}{\textbf{Social-IQ 2.0}} &
\multicolumn{7}{c}{\textbf{YouTube}} \\
\cmidrule(lr){2-8}\cmidrule(lr){9-15}
& \textbf{Avg} & \textbf{P} & \textbf{I} & \textbf{V} & \textbf{O} & \textbf{T} & \textbf{S}
& \textbf{Avg} & \textbf{P} & \textbf{I} & \textbf{V} & \textbf{O} & \textbf{T} & \textbf{S} \\
\midrule

% Separate Prediction & 34.54 & \textbf{62.62} & 23.82 & 33.82 & 23.77 & \textbf{30.28} & 31.50 & \textbf{40.81} & \textbf{67.92} & 32.38 & \textbf{34.91} & \textbf{29.47} & \textbf{35.85} & \textbf{43.14} \\
% \makecell[l]{Separate Prediction \\ with CoT Prompting} & \textbf{41.23} & 60.06 & \textbf{39.45} & \textbf{46.70} & \textbf{26.82} & 26.13 & \textbf{48.15} & 31.48 & 37.96 & \textbf{44.44} & 30.56 & 22.22 & 16.67 & 37.04 \\

Baseline & 34.54 & 62.62 & 23.82 & 33.82 & 23.77 & 30.28 & 31.50 & 40.81 & 67.92 & 32.38 & 34.91 & 29.47 & 35.85 & 43.14 \\
Multi-Stage Prompt & 41.23 & 60.06 & 39.45 & 46.70 & 26.82 & 26.13 & 48.15 & 31.48 & 37.96 & 44.44 & 30.56 & 22.22 & 16.67 & 37.04 \\
In-Context Prompt & 40.51 & 58.96 & 43.39 & 41.95 & 25.41 & 31.06 & 41.88 & 32.40 & 60.75 & 42.99 & 34.26 & 12.96 &  16.82 & 26.85 \\

\bottomrule
\end{tabular}
\caption{\textit{Comparison of prompting strategies for interpersonal relationship scoring.} We evaluate multi-stage and in-context prompting against the separate-prediction baseline, analyzing their effects on PIVOTS dimension scoring performance.}
\label{tab:prompt}
\end{table*}

%% file: latex/tables/related_work.tex
\begin{table*}[!t]
    \centering
    \resizebox{\linewidth}{!}{
    \begin{tabular}{l c c c c}  % 调整列格式为左对齐+居中，更紧凑
        \toprule  
        \textbf{Benchmark} & \textbf{Scale} & \textbf{Core Task} & \textbf{Data Source} & \textbf{Annotation} \\
        \midrule  

        SOCIAL GENOME & 272 videos 1486 QA & Answer/Reasoning & Social-IQ 2.0 & Reasoning, entity, sentiment \\
        \midrule
        
        SOCIALEVAL & 2493 QA & Goal/Interpersonal & Human + Trans. & Social, goal \\
        \midrule
        
        SIV-Bench & 2792 videos 8728 QA & Scene/State/Prediction & TikTok, YouTube & Rel. types, Subtitle \\
        \midrule
        
        SocialMaze & 70K (no video) & Role/Decision/Graph & Synthetic + Real + Human & Role, rel., decision \\
        \midrule

        SCRIPTS & 1,147 dialogues & Social relation reasoning & U.S. and Korean movie scripts & Soft relation labels, relational attributes \\
        \midrule
        
        PIVOTS-Bench (Ours) & 191 videos 765 QA & Rel. scoring/Keyframe/Causal & Social-IQ 2.0, YouTube & 6D score, Key time, Visual cue \\
        \bottomrule 
    \end{tabular}
    }
    \caption{Detailed comparison of benchmark datasets}
    \label{tab:related_work}
\end{table*}

%% file: latex/tables/per_axis.tex
% Requires \usepackage{booktabs}
\begin{table}[t]
\centering
\small
\setlength{\tabcolsep}{4pt}
\renewcommand{\arraystretch}{1.15}
\begin{tabular}{lcc}
\toprule
\textbf{Dimension} & \textbf{Exact Agreement (\%)} & \textbf{Krippendorff's $\alpha$ } \\
\midrule
P & 86.23\% & 0.6434 \\
I & 83.48\% & 0.8402 \\
V & 81.98\% & 0.7693 \\
O & 79.97\% & 0.7803 \\
T & 87.59\% & 0.8395 \\
S & 83.96\% & 0.7906 \\
\midrule
\textbf{Overall} & \textbf{83.87\%} & \textbf{0.8125} \\
\bottomrule
\end{tabular}
\caption{Exact agreement rates and Krippendorff's $\alpha$ per dimension.}
\label{tab:agreement_summary}
\end{table}

%% file: latex/tables/score_distribution.tex
% \definecolor{oursblue}{RGB}{210,240,255}

% \begin{table}[t]
% \centering
% \small
% \setlength{\tabcolsep}{6pt}
% \renewcommand{\arraystretch}{1.15}
% \begin{tabular}{lccccc}
% \toprule
% \textbf{dimension} & \textbf{2} & \textbf{1} & \textbf{0} & \textbf{-1} & \textbf{-2} \\
% \midrule
% P & 13  & 89  & 544 & 104 & 15  \\
% I & 54  & 123 & 105 & 345 & 138 \\
% V & 24  & 86  & 147 & 424 & 84  \\
% O & 96  & 64  & 87  & 195 & 323 \\
% T & 121 & 34  & 229 & 238 & 143 \\
% S & 13  & 73  & 124 & 425 & 130 \\
% \bottomrule
% \end{tabular}
% \caption{Scoring distribution across six dimensions}
% \label{tab:pivots_summary}
% \end{table}

\newcommand{\barrule}[1]{\textcolor{oursblue!80}{\rule{#1em}{1.5ex}}} % 定义画条命令

\begin{table}[t]
\centering
\small
\setlength{\tabcolsep}{3pt}
\renewcommand{\arraystretch}{1.2}
\begin{tabular}{lcr cr cr cr cr}
\toprule
\textbf{Dim.} & \multicolumn{2}{c}{\textbf{-2}} & \multicolumn{2}{c}{\textbf{-1}} & \multicolumn{2}{c}{\textbf{0}} & \multicolumn{2}{c}{\textbf{+1}} & \multicolumn{2}{c}{\textbf{+2}} \\
\midrule
P & 13 & \barrule{0.1} & 89 & \barrule{0.4} & \textbf{544} & \barrule{2.5} & 104 & \barrule{0.5} & 15 & \barrule{0.1} \\
I & 54 & \barrule{0.2} & 123 & \barrule{0.6} & 105 & \barrule{0.5} & \textbf{345} & \barrule{1.6} & 138 & \barrule{0.7} \\
V & 24 & \barrule{0.1} & 86 & \barrule{0.4} & 147 & \barrule{0.7} & \textbf{424} & \barrule{2.0} & 84 & \barrule{0.4} \\
O & 96 & \barrule{0.4} & 64 & \barrule{0.3} & 87 & \barrule{0.4} & 195 & \barrule{0.9} & \textbf{323} & \barrule{1.5} \\
T & 121 & \barrule{0.6} & 34 & \barrule{0.2} & 229 & \barrule{1.1} & \textbf{238} & \barrule{1.1} & 143 & \barrule{0.7} \\
S & 13 & \barrule{0.1} & 73 & \barrule{0.3} & 124 & \barrule{0.6} & \textbf{425} & \barrule{2.0} & 130 & \barrule{0.6} \\
\bottomrule
\end{tabular}
\caption{Scoring distribution across six dimensions}
\label{tab:pivots_summary}
\end{table}

%% file: latex/tables/var.tex
% \definecolor{oursblue}{RGB}{210,240,255}
% \begin{table}[t]
% \centering
% \small
% \setlength{\tabcolsep}{1pt} 
% \renewcommand{\arraystretch}{1.15}

% \begin{tabular}{lccccccc}
% \toprule
% \textbf{Method} & \textbf{Avg} & \textbf{P} & \textbf{I} & \textbf{V} & \textbf{O} & \textbf{T} & \textbf{S} \\
% \midrule

% Qwen3-32B-vl & 0.0468 & 0.3005 & 0.0880 & 0.0860 & 0.0514 & 0.1862 & 0.1352 \\

% \bottomrule
% \end{tabular}
% \caption{The variance of evaluation results over three runs on the YouTube subset.}
% \label{tab:var}
% \end{table}
\begin{table}[t]
\centering
\small
\renewcommand{\arraystretch}{1.15} 
\begin{tabularx}{1.0\linewidth}{lXXXXXX}
\toprule
\textbf{Avg} & \textbf{P} & \textbf{I} & \textbf{V} & \textbf{O} & \textbf{T} & \textbf{S} \\
\midrule
0.0468 & 0.3005 & 0.0880 & 0.0860 & 0.0514 & 0.1862 & 0.1352 \\
\bottomrule
\end{tabularx}
\caption{The variance of evaluation results over three runs on the YouTube subset.}
\label{tab:var}
\end{table}